%% file: main.tex
\documentclass[default,iicol]{sn-jnl}

\usepackage[accsupp]{axessibility} 
\usepackage{graphicx}
\usepackage{amsmath}
\usepackage{amssymb}
\usepackage{xfrac}
\usepackage{booktabs}
\usepackage{multirow}
\usepackage{gensymb}

\newcommand{\ie}{\textit{i}.\textit{e}., }
\newcommand{\eg}{\textit{e}.\textit{g}.\ }
\newsavebox\CBox
\def\textBF#1{\sbox\CBox{#1}\resizebox{\wd\CBox}{\ht\CBox}{\textbf{#1}}}

\newcommand{\F}[1]{\textBF{#1}}
\renewcommand{\S}[1]{\underline{#1}}

\usepackage[capitalize]{cleveref}
\crefname{section}{Sec.}{Secs.}
\Crefname{section}{Section}{Sections}
\Crefname{table}{Table}{Tables}
\crefname{table}{Tab.}{Tabs.}

\jyear{2023}%
\theoremstyle{thmstyleone}%
\theoremstyle{thmstyletwo}%

\theoremstyle{thmstylethree}%
\begin{document}

\input{tex/title}
\input{tex/introduction}
\input{tex/related_works}
\input{tex/method}
\input{tex/experiments}
\input{tex/conclusion}
\input{tex/acknowledgments}



\bibliographystyle{splncs04}
\bibliography{references}

\end{document}

%% file: tex/title.tex
\title[Deep PatchMatch MVS with Learned Patch Coplanarity, Geometric Consistency and Adaptive Pixel Sampling]{Deep PatchMatch MVS with Learned Patch Coplanarity, Geometric Consistency and Adaptive Pixel Sampling}

\author*[1]{\fnm{Jae Yong} \sur{Lee}}\email{lee896@illinois.edu}

\author[2]{\fnm{Chuhang} \sur{Zou}}\email{zouchuha@amazon.com}

\author[1]{\fnm{Derek} \sur{Hoiem}}\email{dhoiem@illinois.edu}

\affil*[1]{
\orgdiv{Department of Computer Science}, \orgname{University of Illinois at Urbana-Champaign}, 
\orgaddress{
\street{201 N Goodwin Ave}, 
\city{Urbana}, 
\postcode{61801}, 
\state{Illinois}, 
\country{United States}}}

\affil[2]{\orgname{Amazon Inc.}, 
\orgaddress{
\street{410 Terry Ave N}, 
\city{Seattle}, 
\postcode{98109}, 
\state{Washington}, \country{United States}}}

\abstract{Recent work in multi-view stereo (MVS) combines learnable photometric scores and regularization with PatchMatch-based optimization to achieve robust pixelwise estimates of depth, normals, and visibility.  However, non-learning based methods still outperform for large scenes with sparse views, in part due to use of geometric consistency constraints and ability to optimize over many views at high resolution.  In this paper, we build on learning-based approaches to improve photometric scores by learning patch coplanarity and encourage geometric consistency by learning a scaled photometric cost that can be combined with reprojection error.  We also propose an adaptive pixel sampling strategy for candidate propagation that reduces memory to enable training on larger resolution with more views and a larger encoder. These modifications lead to 6-15\% gains in accuracy and  completeness on the challenging ETH3D benchmark, resulting in higher $F_1$ performance than the widely used state-of-the-art non-learning approaches ACMM and ACMP.}

\keywords{Multi-View Stereo, Learning-based Stereo, 3D Reconstruction}
\maketitle

%% file: tex/introduction.tex
\input{figures/teasure}

\section{Introduction}
%
Multi-view stereo (MVS) aims to infer accurate and complete 3D geometry from a set of calibrated images, with many applications such as robotics~\cite{furukawa2009accurate,rebecq2018emvs}, mixed reality~\cite{yang2013image,prokopetc2019towards}, and vision-based inspection~\cite{golparvar2011monitoring}.
The first deep learning-based approaches work well for densely sampled views with small-scaled scenes, such as the DTU dataset~\cite{aanaes2016large} and the Tanks-and-Temples intermediate benchmark~\cite{Knapitsch2017tanks}.  However, in more challenging benchmarks with sparse views, wide baselines,  and large depth ranges, such as the ETH3D High-res benchmark~\cite{schoeps2017eth3d}, non-learning based methods achieve better performance by jointly optimizing over many views for the depths, normals, and visibilities of each pixel that maximize photometric scores while satisfying geometric consistency.
Recent methods incorporate PatchMatch optimization into an end-to-end training framework~\cite{wang2020patchmatchnet,lee2021patchmatchrl}, partially closing the performance gap. The latest work PatchMatch-RL~\cite{lee2021patchmatchrl} proposes to jointly learn from pixel-wise depth, normal, and visibility, by using reinforcement learning to overcome the non-differentiable PatchMatch optimization. However, methods such as ACMM~\cite{xu2019multi} still outperform, in part due to the precision of the bilaterally weighted NCC photometric cost, the ability to optimize over many views at high resolution, and the ability to incorporate geometric consistency constraints.

In this paper, we propose three designs to address the above-mentioned issues of learning-based PatchMatch MVS.  
First, we improve the photometric cost function by directly learning whether nearby pixels are co-planar.  Patch-based MVS methods compute photometric score by comparing intensities or features over some neighborhood after rectifying for depth and normal, under the assumption that the surface is locally planar.  Existing approaches weight the neighborhood based on color and pixel distance~\cite{schoenberger2016mvs} or dot-product attention between feature vectors~\cite{lee2021patchmatchrl}. We modify the affinity neighborhood network CSPN~\cite{cheng2019learning} to directly learn which neighboring points are likely coplanar, which provides weights for more robust photometric costs in both textureless and textured regions. 

Second, we incorporate geometric consistency regularization, replacing the recurrent cost regularization of PatchMatch-RL~\cite{lee2021patchmatchrl}. Although geometric consistency is commonly used in non-learning approaches (e.g.,~\cite{schoenberger2016mvs,xu2019multi,Xu2020ACMP}), integration in deep learning approaches is challenging, since the neighborhood geometry is not differentiable with respect to the reference view geometry. 
We side-step this problem by learning to map the photometric score into a log-disbelief (i.e., negative log likelihood) and model reprojection error with a Gaussian distribution, such that the joint likelihood is the sum of photometric log-disbelief and the L2-norm reprojection error. 


Finally, we investigate how to train learning-based PatchMatch MVS to perform inference with high-resolution images, more views, and deeper features. The challenge is that higher resolutions require larger receptive fields to include sufficiently discriminate texture. Coarse-to-fine architectures~\cite{xu2019multi,wang2020patchmatchnet,lee2021patchmatchrl} help, but more complex features with larger receptive fields are also required, which in turn requires prohibitive quantities of GPU memory.  
We propose adaptive sampling of the supporting pixels and gradient checkpointing to reduce the memory usage in the training. This enables more PatchMatch iterations and larger sets of view selections and source views during training, as well as use of a U-Net~\cite{ronneberger2015u}-shaped feature extractor that covers a much larger receptive field compared to the shallow VGG-style backbone used in~\cite{lee2021patchmatchrl}.


In summary, our main \textbf{contributions} are:
\begin{itemize}
    \item \textbf{Learned coplanarity} which weights the photometric score support based on estimated coplanarity.
    \item  \textbf{Modeling geometry consistency guided inference} with learned photometric score functions which improves completion of the reconstruction.
    \item \textbf{Efficient training using adaptively sampled subset of supporting pixels} that reduces memory usage in the training time to allow: a) more PatchMatch training iterations, b) a larger set of source images, c) a larger number of pixel-wise view selections and d) a more complex feature extraction backbone. 
\end{itemize}

Combined, our method outperforms state-of-the-art learning-based MVS methods on the ETH3D high-res benchmark by a large margin with the same image resolution. 
We also provide an efficient and scalable inference architecture that enables high-resolution inference on up to 10 source images of $3200 \times 2112$, outperforming the classical counterpart methods ACMM~\cite{xu2019multi} and ACMP~\cite{Xu2020ACMP} at the same image resolution. 

%% file: figures/teasure.tex
\begin{figure*}[t]
    \centering
    \begin{tabular}{@{}c@{\hskip 0.05in}c@{\hskip 0.05in}c@{\hskip 0.05in}c@{}}
    \includegraphics[width=0.24\textwidth]{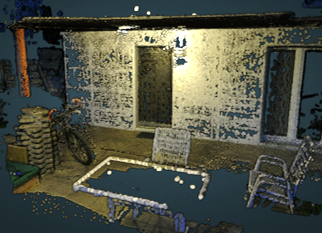} &
    \includegraphics[width=0.24\textwidth]{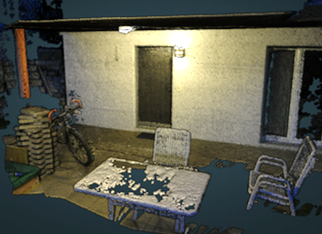} &
    \includegraphics[width=0.24\textwidth]{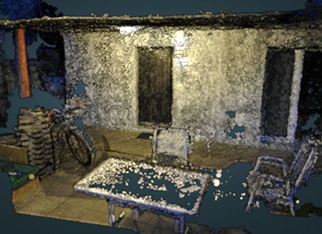} &
    \includegraphics[width=0.24\textwidth]{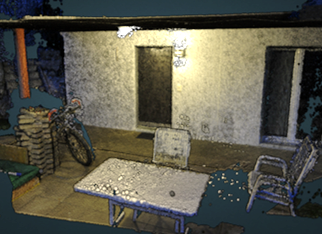} \\
    COLMAP~\cite{schoenberger2016mvs} & ACMM~\cite{xu2019multi} & PatchMatch-RL~\cite{lee2021patchmatchrl} & Ours \\
    \end{tabular}
    
    \caption{
        We propose a learning-based PatchMatch MVS approach with learned patch coplanarity, geometric consistency and adaptive pixel sampling.
        Our method achieves more complete reconstruction compared to existing MVS methods.
    }
    \label{fig:teasure}
\end{figure*}

%% file: tex/related_works.tex
\section{Related Works}
The goal of multi-view stereo (MVS) is to estimate the scene geometry using the images with the known poses. We refer readers to Furukawa and Hern\'{a}ndez~\cite{furukawa2015_mvsTutorial} for a broad review and focus on the most related work here.


Stereo based depth estimation methods hypothesize depth values and score the hypotheses based on photometric similarity (or matching cost).  
Initial works~\cite{lucasKanade1981_stereo,hannah1974computer,anandan1989computational} compute the matching cost as the sum of square differences or absolute differences between corresponding pixels in a local patch. This simple score is unreliable for oblique surfaces, near depth discontinuities, and for low-texture or non-Lambertian surfaces.  
MVS approaches evolved to address these concerns by jointly estimating depth and normal for each pixel~\cite{bleyer2011patchmatch}, assigning weights to supporting pixels~\cite{schoenberger2016mvs}, and using robust matching functions, e.g. learned or NCC, as well as regularizing the cost based on neighborhood estimates or scores~\cite{yao2018mvsnet,yao2019recurrent,gu2020cascade,luo2020attention}. 
The joint depth and normal search has been implemented by using PatchMatch~\cite{barnes2009patchmatch} based optimization by Bleyer et al.~\cite{bleyer2011patchmatch}, and extended to multi-view stereo with visibility selection~\cite{zheng2014patchmatch,galliani2015massively,schoenberger2016mvs}. 
Incorporation of normal estimates enables a more accurate patch-wise homography to match views of oblique surfaces, but depth discontinuities remain a problem.  To address depth discontinuities, bilaterally weighted normalized cross correlation (NCC) is proposed~\cite{schoenberger2016mvs}, with weights modeling the likelihood of the coplanarity of each supporting pixel as proportional to a product of color and position difference terms. Bilaterally weighted NCC is widely used in recent non-learning PatchMatch-based methods~\cite{xu2019multi,romanoni2019tapa,Xu2020ACMP,kuhn2020deepc,wang2020mesh,xu2020marmvs}. 

In learning-based PatchMatch MVS, Lee et al.~\cite{lee2021patchmatchrl} use scaled dot-product based attention to model supporting pixel weights.
Both bilaterally weighted NCC and scaled dot-product attention rely on center-pixel similarity to measure the local co-planarity of the supporting pixels. However, when pixels across depth boundaries have similar colors or flat surfaces have varying texture, center-pixel similarity does not correctly model coplanarity. Instead, we propose to directly regress the supporting pixel-wise co-planarity using ground-truth supervision.

Local photometric scores are unpredictive or unreliable for textureless or non-Lambertian surfaces, requiring consistency checks across neighborhoods or other views.
COLMAP~\cite{schoenberger2016mvs} checks the consistency of predicted depths and normals with neighboring views to augment the photometric score and filter inconsistent points, improving accuracy but decreasing recall. ACMM~\cite{xu2019multi} propose a coarse-to-fine geometric consistency regularized inference. Reconstruction at coarse resolutions enables photometric scores with wide receptive fields with more texture, yielding more accurate but imprecise depth estimates, that are then refined at finer scales using a combination of geometric consistency and photometric cost terms.  
Our work combines the coarse-to-fine geometric consistency driven inference with the learning based PatchMatch MVS by mapping the learned photometric score into a log-disbelief scale, which allows the optional aggregation of additional log-disbelief terms, such as a geometric consistency term. 

DeepC-MVS~\cite{kuhn2020deepc} achieves state-of-the-art results on ETH3D by adding a post-process refinement step to a slightly modified version of ACMM, filtering or refining depth values using learned confidences based on the initial estimates of depth, normal, and visibility.  Similar refinements would likely benefit learning-based approaches also, but we leave this to future work.

The state-of-the-art learning-based methods, Vis-MVSNet~\cite{zhang2020visibility} and EPP-MVSNet~\cite{ma2021eppmvsnet} operate on high-resolution images, which allows more definite reconstruction as pixel-level precision improves as frustums corresponding to each projection pixels become narrower. 
Both Vis-MVSNet~\cite{zhang2020visibility} and EPP-MVSNet~\cite{ma2021eppmvsnet} use U-Net~\cite{ronneberger2015u} as a shared CNN based encoder, which is known to preserve high-frequency details and provide wider receptive fields that make the encoder more suitable for high-resolution reconstruction. 

We additionally integrate the learned patch coplanarity to sample the subset of supporting pixels in training time. This allows time and memory efficient training using policy gradient~\cite{lee2021patchmatchrl}, which lets our method to be trained with more PatchMatch iterations and source view selections per pixel, as well as using 2D U-Net~\cite{ronneberger2015u} based feature extraction.

%% file: tex/method.tex
\input{figures/architecture}
\section{Method}
Figure~\ref{fig:architecture} shows an overview of our architecture during training. We build our approach on PatchMatch-RL~(Sec.~\ref{method:PMRL}), a recent learning-based PatchMatch MVS framework, and propose three designs that lead to large performance gain in both accuracy and completeness: (1) improve the photometric cost function by directly modeling the coplanarity of neighboring pixels~(Sec.~\ref{method:coplanarity}); (2) replace the recurrent cost regularization with geometric consistency by learning a scaled photometric cost ~(Sec.~\ref{method:gc}); (3) design an adaptive pixel sampling strategy for candidate propagation to reduce memory, so that high-res images, more views, and deeper features can be used for training~(Sec.~\ref{method:adaptive}).


\input{figures/inference}

\subsection{Review of PatchMatch-RL}\label{method:PMRL}
The input of PatchMatch-RL is a set of reference and source images $\{I_r, I_i\}^N_{i=1} \subset \mathcal{I}$, and their corresponding cameras poses $\{C_r, C_i = (K_i, R_i, t_i)\}^N_{i=1} \subset \mathcal{C}$ including intrinsic and extrinsic parameters. 
The method uses a coarse-to-fine PatchMatch MVS inference to jointly estimate the depths, normals, and visiblities of the reference image $D_r, N_r, \{V_{r,i}\}^N_{i=1}$, and uses policy-gradient for end-to-end training.

PatchMatch-RL starts by extracting CNN based features $\{f^c_{\theta}(I_r)=F^s_r, f^c_{\theta}(I_i) = F^{s}_i\}^N_{i=1}$ at multiple scales $s \in \{3, 2, 1\}$, where $f^c_{\theta}$ represents trainable CNN backbone with parameters $\theta$, and $s$ represents the $\log_2$ scale resolution of the image (\eg $F^s_i \rightarrow \mathbb{R}^{\frac{H}{2^s} \times \frac{W}{2^s} \times C}$). Then, the method initializes the depth and normal map $d^{s,t}_r,n^{s,t}_r$ at the coarsest scale $s=3$ and the initial iteration $t=0$.
For the ease of notation, we drop the scale $s$ and let $d^t_r, n^t_r, v^t_{r,i}$ denote the depth, normal and visibility map at the $t$-th iteration on the same scale in the coarse-to-fine stage from onward. 

After initialization, a series of PatchMatch iterations are run at each scale, and each iteration jointly updates the depth, normal, and visibility maps. A single PatchMatch iteration consists of 4 steps: 
\begin{enumerate}
    \item \textbf{Pixel-wise view selection} updates the visiblity map $v^{t}_{r, i} \rightarrow v^{t+1}_{r, i}$ using the current set of depth $d^{t}_r$ and normal $n^{t}_r$ scored by an MLP $f^v_\theta$, which defines the parameterized view-sampling policy $\pi^v_\theta$.
    \item \textbf{Candidate Propagation} generates potential candidates of depth and normal for the next iteration by using PatchMatch-based propagation. 
    \item \textbf{Candidate Scoring} scores each candidate depth / normal pairs using the current set of visibility map $v^{t+1}_{r, i}$ through an RNN-based scorer regularizer $f^s_\theta$, which defines the parameterized candidate-sampling policy $\pi^s_\theta$.
    \item \textbf{Candidate Selection} updates the current depth and normal map to $d^{t+1}_r$ and $n^{t+1}_r$,
\end{enumerate}

The parameterized policies represents the sampling procedure defined by the distribution formed by the scores obtained by the MLP $f^v_\theta$ and the RNN $f^s_\theta$. 
The view selection score distribution takes the supporting pixel-aggregated group-wise correlation scores for each source images as input. 
The supporting pixel-aggregated group-wise correlation scores use supporting pixel weights, obtained by the dot-product attention between the features at the center pixel and the supporting pixels, to compute weighted mean of the group-wise correlation scores between the pixels of reference feature map and the corresponding pixels of the source feature map. 
Given supporting pixels $q \in W(p)$ around the pixel $p$ and its corresponding weights $w(q)$, the supporting pixel-aggregated group-wise correlation of the pixel $p$ can be defined as:
\begin{align}
    \mathcal{G}(p) = \sum_{s=1}^N \sum_{q \in W(p)} w(q) (F^r(q) \circledast F^s(q')) / N
    \label{eq:gc}
\end{align}
where $q'$ indicates the corresponding pixel of $q$ in the source image warped by the depth $d_r(p)$ and normal $n_r(p)$ value at $p$.

Similarly, the candidate scoring distribution obtained through $f^s_\theta$, takes view-weighted, supporting pixel aggregated group-wise correlation scores for selected views as input:
\begin{align}
    \mathcal{G}^{\mathcal{V}}_{\phi}(p) = \sum_{s \in \mathcal{V}} \sum_{q \in W(p)} v_{r,s}(p) w(q) (F^r(q) \circledast F^s(q^\phi))
    \label{eq:wgc}
\end{align}
where $\mathcal{V}$ is the set of selected views, and $v_{r,s}(p)$ indicates visibility estimate of the pixel $p$ in the reference image $r$ from the source image $s$, and $q^\phi$ denotes the corresponding pixel of $q$ in the source image, warped by the candidate depth $d_\phi(p)$ and normal $n_\phi(p)$. 
The output value of the RNN $f^s_\theta$ measures the photometric score of each candidate as $S^\phi_{pho}(p) = f^s_\theta(\mathcal{G}^{\mathcal{V}}_\phi(p))$. 

In inference, the policy performs arg-max selection based on the set of computed scores. In training, the policy performs selection based on the decaying $\epsilon$-greedy method, where it samples from the distribution with the probability of $\epsilon$ and samples using arg-max with the probability of $1-\epsilon$. 

After a sufficient number of iterations for each scale, the depth and normal maps are upscaled using nearest neighbor upsampling. Then, the next stage of the coarse-to-fine architecture starts, and the process is repeated until the most finest-level iteration is completed.

\input{figures/coplanarity}
\input{figures/coplanarity_comparison}

To train the model in an end-to-end differentiable manner, PatchMatch-RL uses policy-gradient algorithm REINFORCE to jointly train $f^c_\theta, f^v_\theta, f^s_\theta$. 
The reward for each iteration is defined as $r^t = \mathcal{N}(d^t_r; D^*_r, \sigma_d) \cdot \mathcal{N}(n^t_r; N^*_r, \sigma_n)$, where $D^*_r$ and $N^*_r$ represent the ground-truth depth and normal maps, and $\mathcal{N}(x; \mu, \sigma)$ represents the probability of observing $x$ from the normal distribution with mean $\mu$ and standard deviation $\sigma$. 
The update for each iteration is:
\begin{align}
    \triangledown_\theta J &=  \sum_t \sum_{t'>=t} \triangledown_\theta \gamma^{t'-t} r^t \log \pi_\theta.
\end{align}
We let $\gamma$ represent the discount coefficient of future reward. PatchMatch-RL additionally modifies the candidate selection loss to use cross-entropy between the ground-truth distribution of observing each candidate (\eg $\forall (d, n), \mathcal{N}(d; D^*_r, \sigma_d) \cdot \mathcal{N}(n; N^*_r, \sigma_n)$) and the distribution formed by the scores of each candidates using the candidate scoring policy for robustness in training. 

\subsection{Learning Patch Coplanarity}\label{method:coplanarity}
Estimating the surface normal for each pixel requires a patch-based photometric score, computed as a weighted sum of similarities over supporting interpolated feature positions.  Rather than computing weights using scaled dot-product attention like PatchMatch-RL, we propose to set the weights to the likelihood of coplanarity with supporting pixels, which is estimated using a separate CNN architecture.
(\ie we directly estimate $w(q)$ for $q \in W(p)$ in Eqs.~\ref{eq:gc} and \ref{eq:wgc})
We modify the CSPN architecture~\cite{cheng2019learning} used to learn the affinity of neighboring points, and change its last layer to a $3\times 3$ convolution with a dilation of 3, so that the receptive field matches the supporting pixel weight locations.
Figure~\ref{fig:coplanar} depicts an overview of our coplanarity extraction and output feature representation. 
In addition to the policy gradient loss, we use mean squared error term from the ground-truth coplanarity points obtained from the ground-truth depth to supervise the the training. 
We show in Figure~\ref{fig:coplanar-comparison} that compared with existing supporting pixel weighting schemes, our approach can better identify coplanarity within local patches with different textures.

\subsection{Encouraging Geometric Consistency in Inference}\label{method:gc}
We propose geometric consistency based regularization 
to replace the recurrent cost regularization used in PatchMatch-RL. The key idea of geometric consistency is to use the inferred geometry of the source images to validate the reference image geometry. We adopt the geometric consistency score used in ACMM~\cite{xu2019multi}, which defines an additional score term based on the reprojection error of the visible source view pixels.

Although geometric consistency is commonly used for non-learning based approaches~\cite{schoenberger2016mvs,xu2019multi}, it is challenging to integrate in learning-based approaches, since the source view geometry is not differentiable with respect to the reference view geometry. 
In order to allow the optional adding of geometric log-disbelief scores in candidate selection stage without having to train additional module to combine the two scores, we remap the photometric score and the geometric consistency score range to the log-disbelief range. 
In detail, we let the photometric score ranges logarithmically from $0$ to $S_{max}$, where $0$ implies perfect photometric match and $S_{max}$ implies complete disagreement.
We use MLP as the photometric consistency scorer to replace the RNN-based scorer $f^s_\theta$, since the history of candidate selection cannot handle the additional geometric consistency term in inference. 

We let the pixel-wise reprojection error distance be the log scale disbelief score of the geometric consistency term.
Given the candidate depth and normal map $d_\phi, n_\phi = \phi$, we use the homography between image $j$ and image $k$ at pixel $p$ defined as $H^\phi_{j \rightarrow k}(p) = K_k \cdot (R_{j\rightarrow k} - \frac{t_{j\rightarrow k} \cdot n^T_\phi(p)}{d_\phi(p)}) \cdot K^{-1}_j$, and compute the reprojection error:
\begin{equation}
    E^\phi_{s}(p) = || H^\phi_{s \rightarrow r}(H^\phi_{r \rightarrow s}(p) \cdot p) -  p ||_2
\end{equation}
We define $G_{max}$ as the maximum log-disbelief term that can be added to geometric consistency, and combine the score weighted by the visibility mapping weights:
\begin{equation}
    S^{\phi}_{geo}(p) = \sum^{N}_{i=1} v^t_{r,i} \cdot (S^{\phi}_{pho}(p) + \min(E^\phi_{i}(p), G_{max}))
\end{equation}
where $S^{\phi}_{pho}(p)$ denotes the photometric score obtained from the MLP.
We then use $S^\phi_{geo}(p)$ to select candidate for the pixel $p$, updating the depth $d_r(p)$ and normal $n_r(p)$ at the corresponding iteration.

Since we need to have the source view geometry available to use geometric consistency based regularization, we use a modified version of the coarse-to-fine architecture, as illustrated in Figure~\ref{fig:inference}.
In inference, the source view and its inferred geometry in the previous stage are used to calculate $S^{\phi}_{geo}$ for the candidate selection step of each PatchMatch iteration. The inferred depth and normal maps at each stage are stored in a file, and loaded in the next stage of reconstruction with geometric consistency.

\subsection{Adaptive Supporting Pixel Sampling}\label{method:adaptive}
Training the learning-based PatchMatch MVS model using policy gradient requires memory linear to the number of depth, normal and visibility map evaluations (\eg number of PatchMatch iterations), since the history of gradients need to be stored. 
Hence, PatchMatch-RL~\cite{lee2021patchmatchrl} uses a shallow VGG-style backbone, with a small number of PatchMatch iterations per scale. 
We resolve this problem by using a sampled subset of the supporting pixels in training, which effectively reduces the training computation by half and the memory by roughly 30\%, so that a larger feature extraction backbone can be used with faster training. 
We use the distribution of the coplanar values of the supporting pixels as a sampling distribution. 
We sample 3 points out of 8 supporting pixels, (excluding the center pixel of the $3\times 3$ local supporting pixels), which is the minimum number of points that supports estimates of surface normal orientation. 
The adaptive supporting pixel sampling can be used in the inference time for faster runtime, at the cost of a slight drop in reconstruction quality. 
Table~\ref{tab:ablation} and Table~\ref{tab:ablation_hr} in Section~\ref{sec:experiments} show the detailed analysis of the effect of adaptive supporting pixel sampling in inference. 

%% file: figures/architecture.tex
\begin{figure*}[t] \centering
    \includegraphics[width=1.0\textwidth]{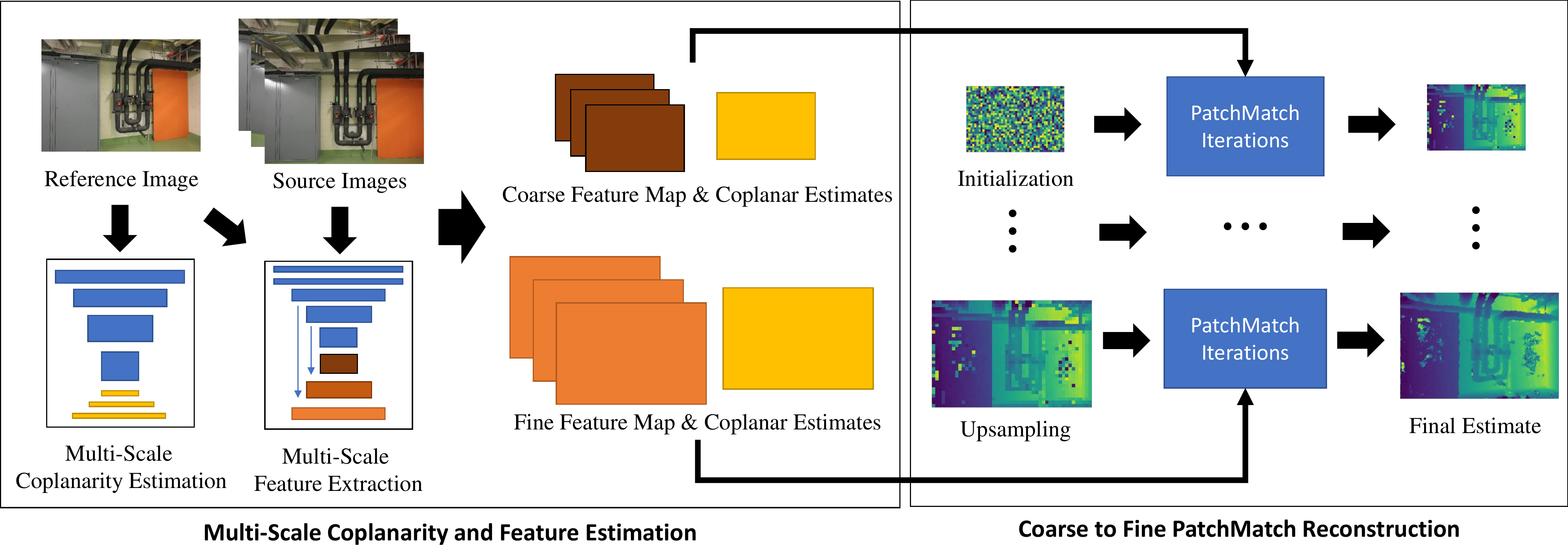}
    \caption{
        \textbf{Training Architecture overview: }
        We extract multiscale features for each image and multiscale coplanarity weights for the reference image. 
        We use features and coplanar maps of corresponding scales to perform coarse-to-fine estimation of depth, normal and visibility maps. 
        At the coarsest stage, we initialize pixelwise depth and normal maps.  Then, a series of PatchMatch iterations update the estimates.
        The PatchMatch iteration consists of four stages: (1) pixelwise view selection; (2) candidate propagation; (3) candidate scoring; and (4) candidate selection.  
        The current solution is then upsampled to a finer level of input through nearest neighbor interpolation, and this procedure continues until the oriented point estimates at the finest level are fused from all images.
    }
    \label{fig:architecture}
 \end{figure*}

%% file: figures/inference.tex
\begin{figure*}[ht] \centering
    \includegraphics[width=1.0\textwidth]{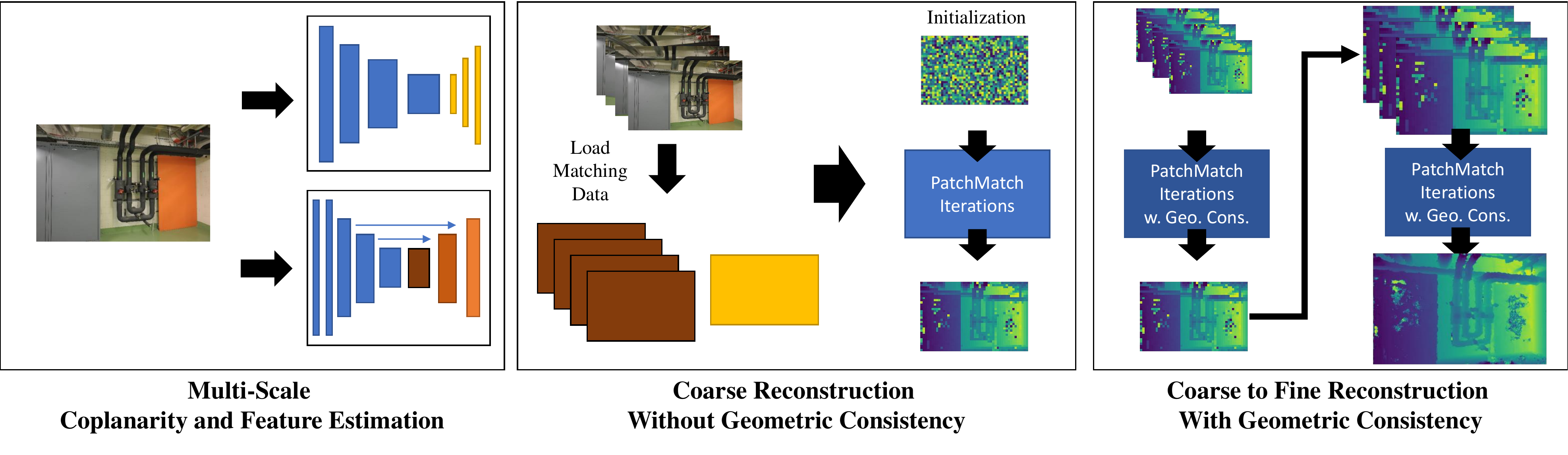}
    \caption{
        \textbf{Inference Architecture: }
        First, features and coplanar maps for each image are generated and stored in files.
        Then, coarse-level reconstruction without geometric consistency  constraints is performed for each reference image.  Specifically, the relevant coarse coplanar and feature maps are loaded and, after initializing depths and normals, a PatchMatch iteration (Figure~\ref{fig:architecture}) is performed to infer geometry. 
        Next, coarse-to-fine reconstruction is performed with geometric consistency. For each stage, results from the previous stage are used to initialize estimates and compute geometric consistency scores.  All images in the scene are processed stagewise, and the inference is complete once the finest scale is processed.
    }
    \label{fig:inference}
 \end{figure*}

%% file: figures/coplanarity.tex
\begin{figure*}[t] \centering
    \includegraphics[width=0.85\textwidth]{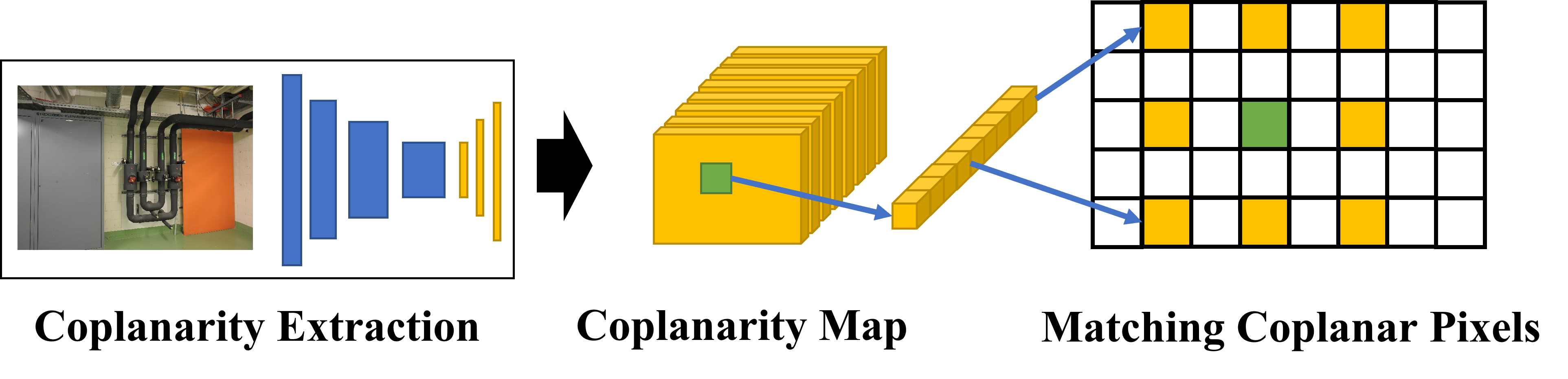}
    \caption{
        \textbf{Coplanarity Extraction and Representation: }
        We use a coplanar extractor to estimate the coplanar values of the supporting pixels of each pixel in the image. 
        The output coplanarity map is an array of shape $H \times W \times 9$, where each position represents the corresponding $3\times 3$ supporting pixels of the image. 
    }
    \label{fig:coplanar}
 \end{figure*}

%% file: figures/coplanarity_comparison.tex
\begin{figure}[t] \centering
    \includegraphics[width=0.475\textwidth]{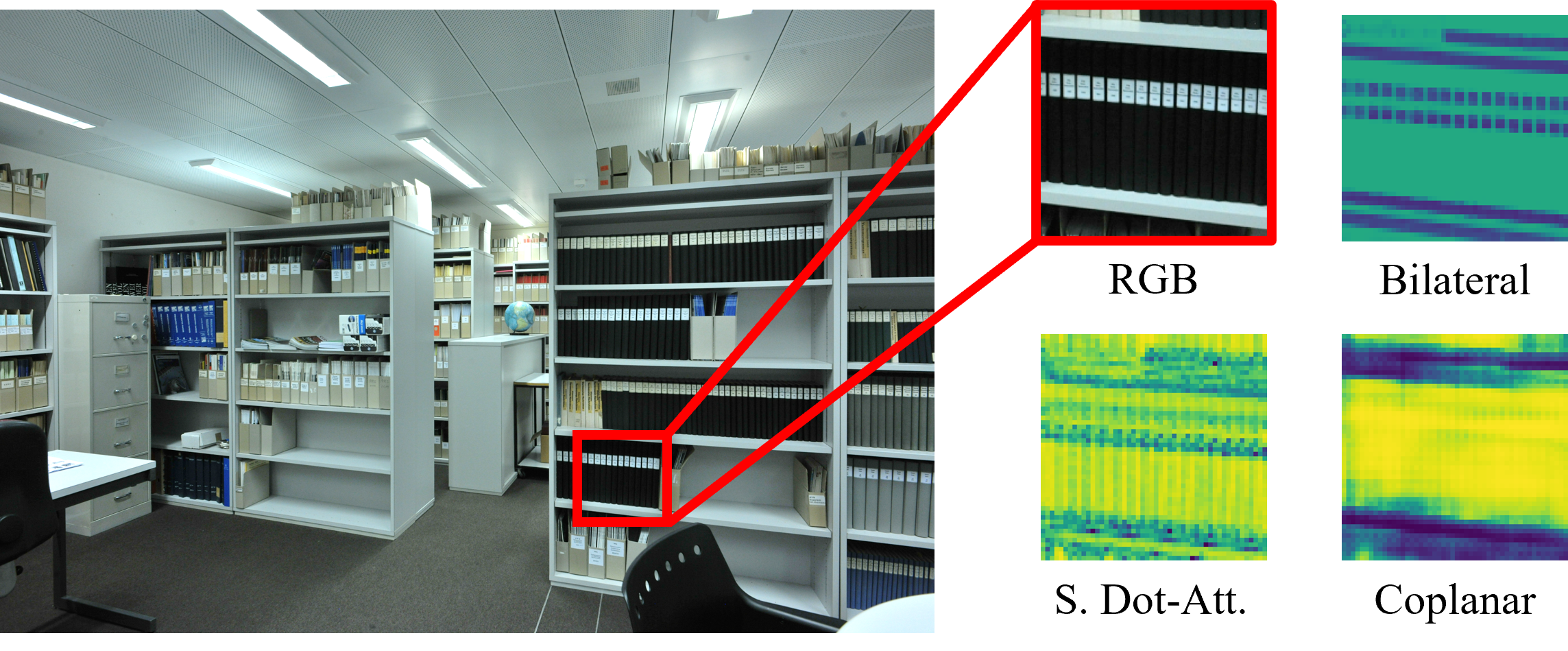}
    \caption{
        \textbf{Comparison of Supporting Weights: }
        We visualize different types of the supporting pixel weights that are 3 pixel above from the center. 
        From the left top, RGB denotes the extracted patch for the reference, Bilateral denotes the bilateral weighting~\cite{schoenberger2016mvs}, S. Dot-Att. denotes scaled dot-product attention~\cite{lee2021patchmatchrl} and Coplanar denotes our learned coplanar weights. 
        Our method more correctly identifies which pixels correspond to coplanar surfaces, e.g. tags on binders are coplanar with dark part of binders, while shelves are not.
    }
    \label{fig:coplanar-comparison}
 \end{figure}

%% file: tex/experiments.tex
\input{figures/qualitative_results}
\section{Experiments}
\label{sec:experiments}
We evaluate our method on two standard MVS benchmarks, ETH3D High-res Multi-View benchmark~\cite{schoeps2017eth3d} and Tanks-and-Temples(TnT)~\cite{Knapitsch2017tanks} intermediate and advanced benchmark. 

\subsection{Implementation Details}
For all benchmarks, we use BlendedMVS~\cite{yao2020blendedmvs} dataset for training. 
BlendedMVS is a large-scale MVS dataset that contains 113 indoor, outdoor and object-level scenes. We use 7 images (\ie 1 reference image and 6 source images) with an input resolution of $768 \times 576$, and an output resolution of $384 \times 288$ for training. 
Among the 6 source images, we use 3 images sampled from the 10 best matching source images given by the global view selection provided by the dataset, and 3 images randomly selected from the remaining images in the scene. 
In training local view selection, we consider the three best views to be visible and the two worst views to be not visible. In inference, we select the three best source views for reconstruction.
Regarding the number of PatchMatch iterations for each scale, we use $N^3_{iter}, N^2_{iter}, N^1_{iter}$ = $8, 2, 1$ in training, and $8, 3, 3$ in inference respectively.
We finally fuse the estimated depth and normal maps using the fusion method of Galliani et al.~\cite{galliani2015massively}, where we filter depths by the number of minimum consistent views containing relative depth error below 1\%, reprojection error less than 2px and normal angle difference less than 10$\degree$. We apply nearest neighbor upsampling with $5\times 5$ median filtering to upscale the depth map to the same resolution of the original image during fusion. 
We implemented our method in PyTorch and use RTX 3090 for training and evaluation. 
We use a customized CUDA kernel to accelerate the computation of supporting pixel-aggregated group-wise correlation in inference. These design choices are evaluated in our ablation study (Sec.~\ref{exp_ablations}, Table~\ref{tab:ablation_additional}).
\input{tables/eth3d}

\subsection{ETH3D High-res Multi-View Benchmark}
ETH3D High-res Multi-View Benchmark is one of the more challenging benchmark in MVS, covering large scale scenes with sparsely sampled, high-resolution images of $6048 \times 4032$. 
The benchmark contains training and testing scenes, where training scenes contain 7 indoor and 6 outdoor scenes, and testing scenes contain 9 indoor scenes and 3 outdoor scenes. 
We evaluate our method for both sets of scenes at two different resolution; higher resolution of $3200 \times 2112$ ($\text{Ours}_{hr}$) which matches the resolution of ACMM~\cite{xu2019multi}, and lower resolution of $1920 \times 1280$ ($\text{Ours}_{lr}$) which matches the resolution of Vis-MVSNet~\cite{zhang2020visibility} and PatchMatch-RL~\cite{lee2021patchmatchrl}. We use 15 source views and 10 source views for $\text{Ours}_{lr}$ and $\text{Ours}_{hr}$ respectively.
\input{figures/pointcloud_results}

Table~\ref{tab:eth3d} shows the quantitative comparison of our method. 
Our method achieves the state-of-the-art result among all the learning-based methods even with using the \textbf{lower-resolution} images, outperforming EPP-MVSNet~\cite{ma2021eppmvsnet} by 4.1\% and 3.5\% in the $F_1$ score for the train set in the 2cm and 5cm threshold, and differing by 0.8\% for the test set in the 5cm threshold. 
Moreover, $\text{Ours}_{hr}$ achieves better $F_1$ score than ACMP~\cite{Xu2020ACMP} and ACMM~\cite{xu2019multi} for all settings. Specifically, our $F_1$ scores are higher by 1.2\% and 2.9\% at 2cm and 5cm thresholds on the training sets and by 4.3\% and 4.4\% at 2cm and 5cm thresholds on the test sets, compared to ACMM~\cite{xu2019multi}.
In Figure~\ref{fig:eth3d}, we compare the depth and normal maps of our method against PatchMatch-RL~\cite{lee2021patchmatchrl}. In Figure~\ref{fig:eth_epp_acmm} we compare the point cloud reconstruction with ACMM~\cite{xu2019multi}, EPP-MVSNet~\cite{ma2021eppmvsnet} and ours. 
We show that our method achieves far less noise and is more complete in challenging texture-less surfaces compared to the baseline. 
\input{tables/tanks_and_temples}
\input{figures/tanks_and_temples}
\input{tables/ablations}
\input{tables/supp/ablation_hr}
\input{tables/supp/ablation_additional}
\subsection{Tanks and Temples Benchmark}
We further evaluate our method on Tanks and Temples~\cite{Knapitsch2017tanks} intermediate and advanced benchmark. 
The intermediate benchmark contains 8 scenes captured inward-facing around the center of the scene, and the advanced benchmark contains 6 scenes covering larger area. 
We use 15 images of resolution $1920\times 1072$ for all benchmarks. 
Table~\ref{tab:tnt} compares related work.

We improve the $F_1$ score by 3.5\% and 2.0\% on Intermediate and Advanced scene compared to  PatchMatch-RL~\cite{lee2021patchmatchrl}. 
Our method achieves slightly worse results on the intermediate scenes and on-par performance on the advanced scenes compared to ACMM~\cite{xu2019multi}, differing by 2.0\% and 0.2\% in $F_1$ score respectively. 
We also note that we are the second best learning based method after EPP-MVSNet~\cite{ma2021eppmvsnet} on the advanced benchmark. 

Figure~\ref{fig:tnt_qual} shows the point cloud reconstruction of our method compared to PatchMatch-RL.  %

\subsection{Ablations}\label{exp_ablations}
\textbf{Ablations of design choices.} We provide an extensive study on how different component affects the overall reconstruction quality by using ETH3D High-Res multi-view benchmark~\cite{schoeps2017eth3d} on the training sets. We show ablation studies at higher resolution 3200 × 2112 ($\text{Ours}_{hr}$) and lower resolution 1920 × 1280 ($\text{Ours}_{lr}$) in Table~\ref{tab:ablation} and Table~\ref{tab:ablation_hr} respectively. For a fair comparison, we use the same image resolution and number of views used as our approach in each table. We show that the ablation on $\text{Ours}_{hr}$ shows similar results to $\text{Ours}_{lr}$. 

\input{figures/supp/jpeg_artifact}

\noindent\textbf{Importance of geometric consistency:} Taking the higher resolution as an example, having Geometric consistency improves the $F_1$ score by 1.6\% at the 2cm threshold, and 1.2\% at the 5cm threshold. Without Geometric consistency, the $F_1$ score drops by 0.7\% at the 2cm threshold and 0.5\% at the 5cm threshold.
However, the overall runtime of the method increases, taking more than 8 seconds on average per image, since geometric consistency requires multi-stage reconstruction. 

\noindent\textbf{Importance of feature extraction backbone:}
We compare the CNN-based feature extractor used in PatchMatch-RL~\cite{lee2021patchmatchrl}, which uses shallow VGG-style layers with FPN connections, to our more complex U-Net~\cite{ronneberger2015u}-based feature extractor. 
From the baseline, at higher resolution, U-Net~\cite{ronneberger2015u} based feature extractor improves the $F_1$ score by 2.3\% and 1.8\% at the 2cm and the 5cm threshold. Without U-Net~\cite{ronneberger2015u} based feature extractor, the $F_1$ score drops by 1.4\% at the 2cm threshold and 1.3\% at the 5cm threshold.

\noindent\textbf{Dot-product attention vs. Learned coplanarity:}
We introduce learned coplanarity for computing weights of each supporting pixels, and compare with the dot-product attention based weighting scheme used in PatchMatch-RL~\cite{lee2021patchmatchrl}. 
Compared to dot-product attention, 
at higher resolution, learned coplanarity improves the overall $F_1$ score by 1.4\% at the 2cm threshold and 1.0\% at the 5cm threshold.
Without the learned coplanarity, the $F_1$ score drops by 1.0\% at both thresholds.

\input{figures/supp/eth3d_hr_train}

\input{figures/supp/eth3d_lr_train}
\input{figures/supp/tnt}

\input{figures/supp/tnt_advanced}

\textbf{Ablations of other engineering tweaks.} We additionally evaluate other factors that contribute to the improvement of our model over PatchMatch-RL~\cite{lee2021patchmatchrl}. We experiment on $\text{Ours}_{lr}$.  We first show that saving bilinearly downsampled images with JPEG format reduces the overall $F_1$ score by 6.5\% at the 2cm threshold and 4.2\% at the 5cm threshold.

This is due to the artifact in JPEG compression as shown in Figure~\ref{fig:downsample_artifact}. 
Next, we show that training with fewer of view selections with fewer iterations (\ie 1 view as visible from 5 source views with 3, 1, 1 iterations, instead of 3 views as visible from 6 source views with 8, 2, 1 iterations) drops the overall $F_1$ score by 1.8\% and 0.8\% at 2cm and 5cm threshold. 
Using upsampling followed by median filtering boosts the $F_1$ score at the 2cm threshold by 3.3\% and by 0.5\% at the 5cm threshold. 

We also evaluate with adaptive sampling at inference time. We show that using 3 supporting pixels selected by adaptive sampling (``Adapt-3'') drops $F_1$ score by 2.4\% and 1.0\% at the 2cm threshold and the 5cm threshold, but reduces 1.5 seconds for time taken per view. With 5 supporting pixels (``Adapt-5''), the $F_1$ drops slightly by 0.3\% at the 2cm threshold with time taken per view reduced by 0.89 seconds per view. 

Finally, we compare the runtime statistics of the model without CUDA kernel code (\ie pure PyTorch~\cite{NEURIPS2019_9015} implementation). The CUDA kernel code reduces per-view inference time by 13.35 seconds and memory by 5,181 MiB.

\subsection{Additional Qualitative Results}
Figure \ref{fig:eth_hr_train} and \ref{fig:eth_lr_train} show our point cloud reconstruction of the ETH3D~\cite{schoeps2017eth3d} High-res train and test set for $\text{Ours}_{hr}$ and $\text{Ours}_{lr}$ respectively.
Figure~\ref{fig:tnt_intermediate} and Figure~\ref{fig:tnt_advanced} show our results for the Tanks and Temples~\cite{Knapitsch2017tanks} dataset.

\subsection{Limitations}

Our method performs well for high resolution sparse views of large scenes in a balanced accuracy and completeness score.  However, non-learning based methods such as ACMM~\cite{xu2019multi} achieve higher accuracy, and cost-volume based deep approaches, such as EPP-MVSNet~\cite{ma2021eppmvsnet}, are better suited to smaller scale scenes with dense views. Also, our inference with geometric-consistency is slower than some other deep MVS algorithms~(\eg \cite{wang2020patchmatchnet}), due to file IO constraints in the multiscale inference.

%% file: figures/qualitative_results.tex
\begin{figure*}[t]
    \centering
    \begin{tabular}{@{}c@{}cc@{}cc@{}c@{}}
        \includegraphics[width=0.16\textwidth]{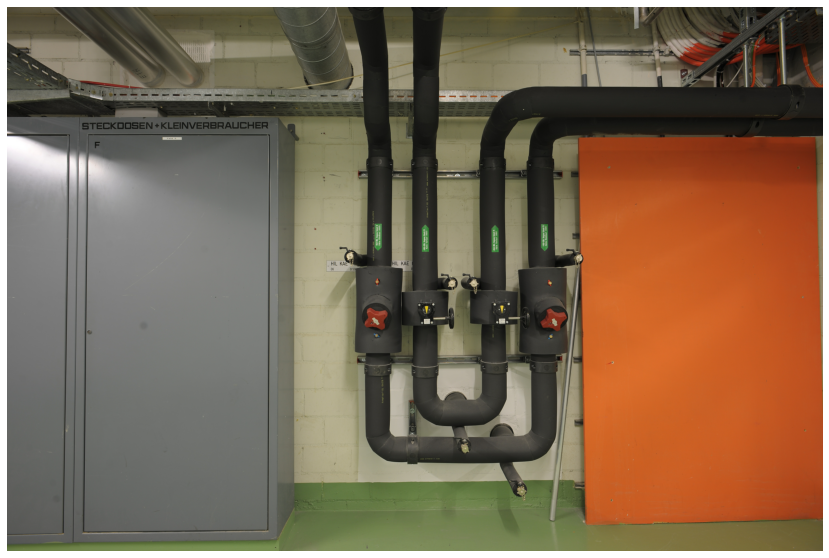} &
        \includegraphics[width=0.16\textwidth]{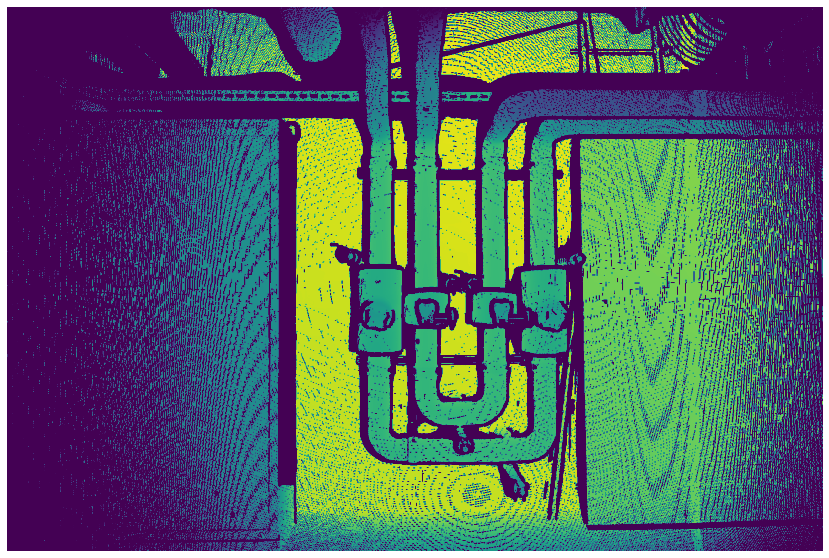} &
        \includegraphics[width=0.16\textwidth]{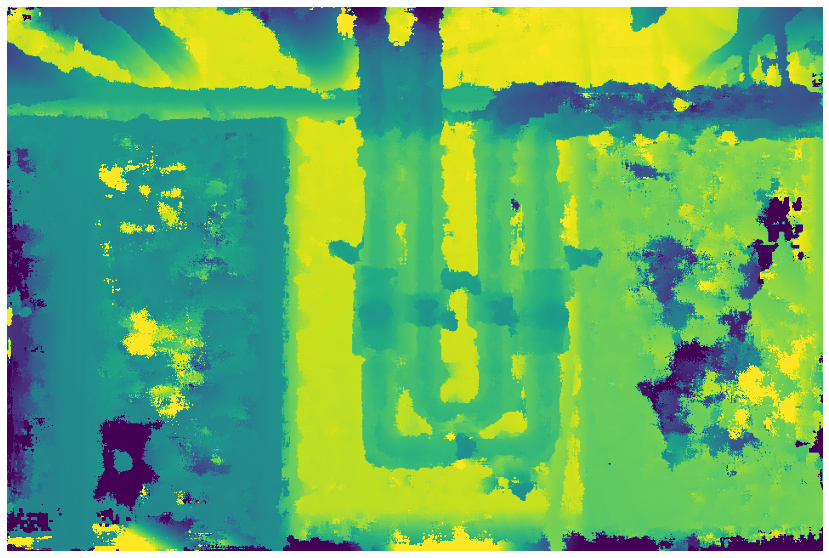} &
        \includegraphics[width=0.16\textwidth]{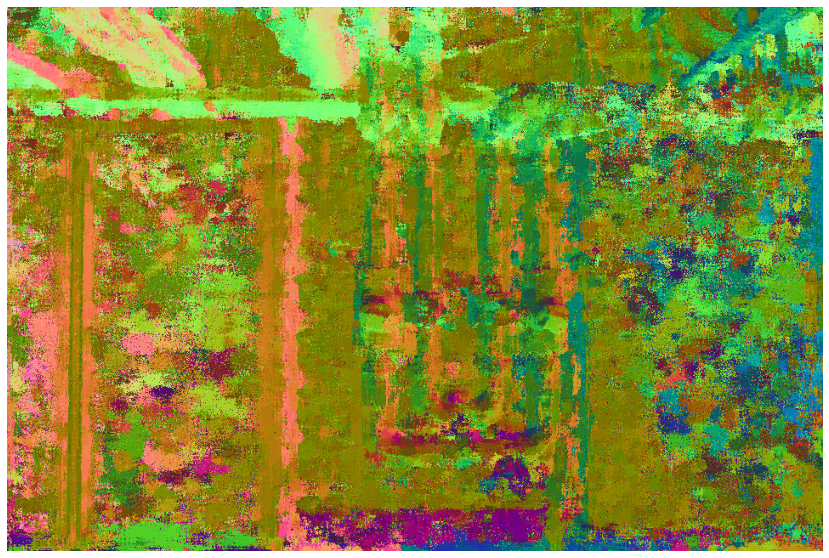} &
        \includegraphics[width=0.16\textwidth]{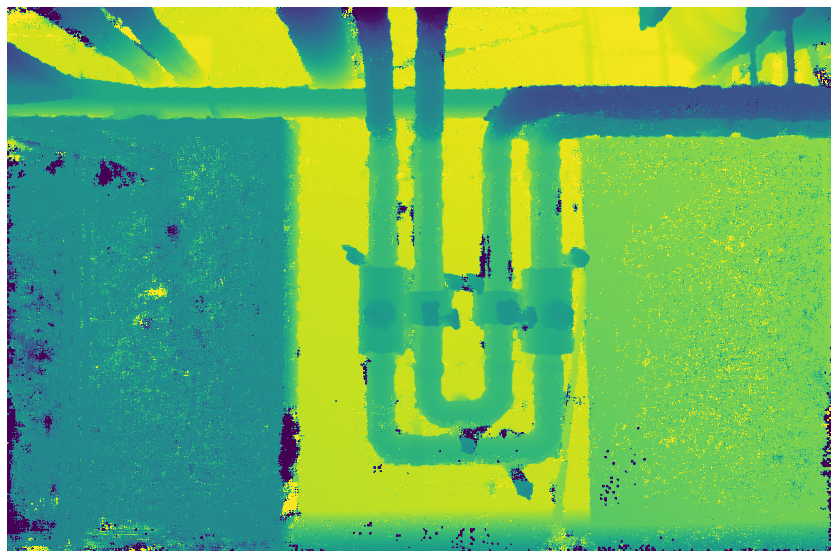} &
        \includegraphics[width=0.16\textwidth]{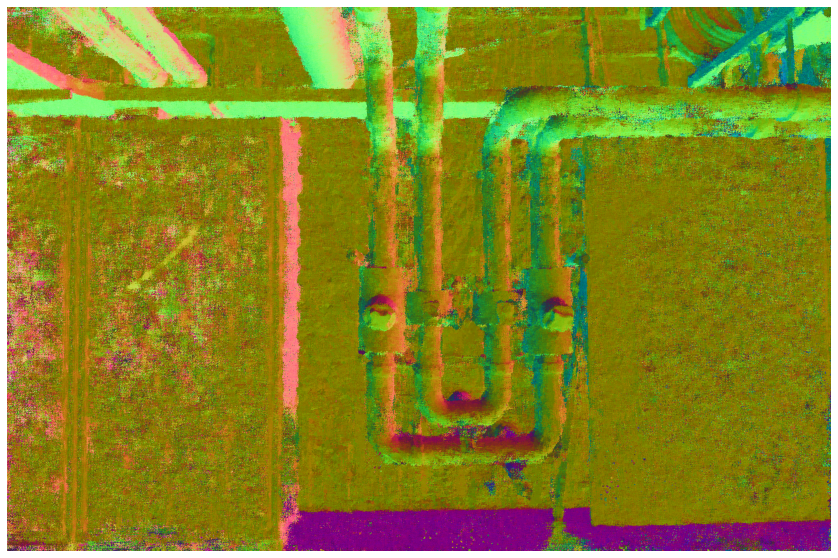}
        \\
        \includegraphics[width=0.16\textwidth]{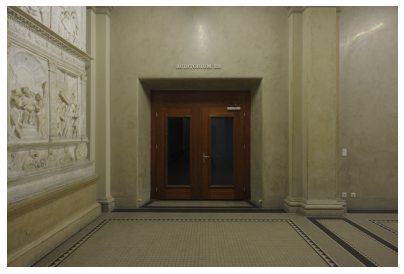} &
        \includegraphics[width=0.16\textwidth]{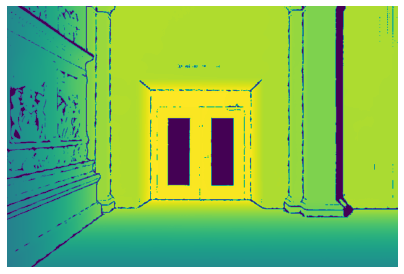} &
        \includegraphics[width=0.16\textwidth]{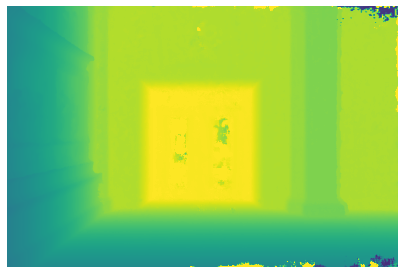} &
        \includegraphics[width=0.16\textwidth]{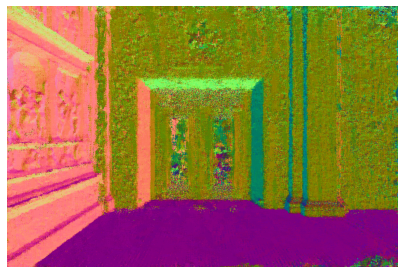} &
        \includegraphics[width=0.16\textwidth]{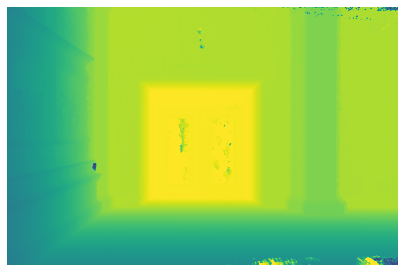} &
        \includegraphics[width=0.16\textwidth]{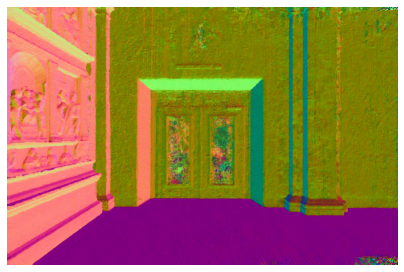}
        \\
        \includegraphics[width=0.16\textwidth]{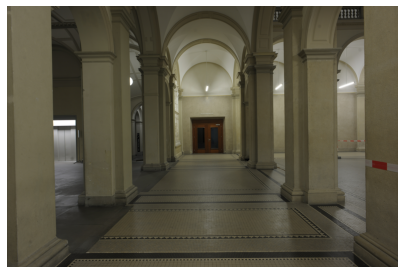} &
        \includegraphics[width=0.16\textwidth]{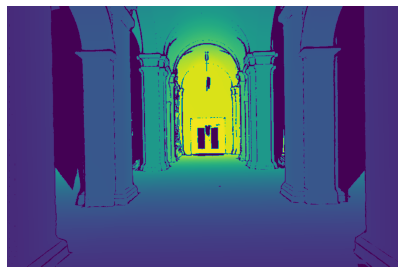} &
        \includegraphics[width=0.16\textwidth]{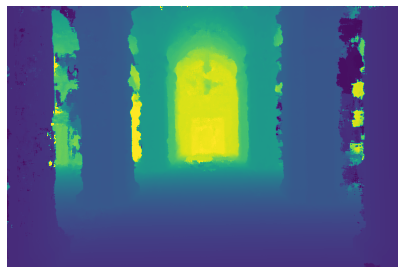} &
        \includegraphics[width=0.16\textwidth]{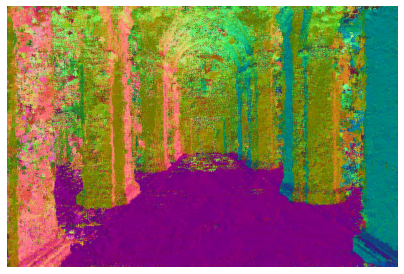} &
        \includegraphics[width=0.16\textwidth]{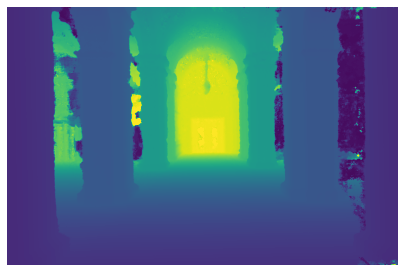} &
        \includegraphics[width=0.16\textwidth]{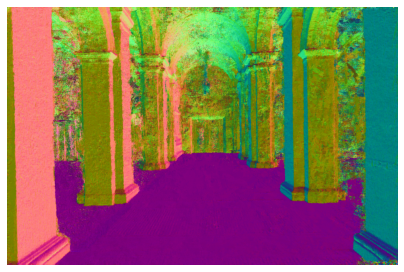}
        \\
        \includegraphics[width=0.16\textwidth]{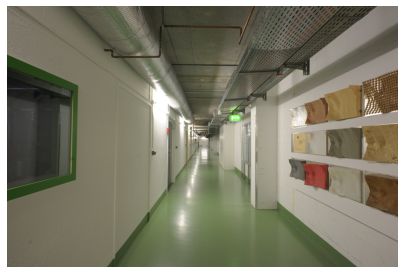} &
        \includegraphics[width=0.16\textwidth]{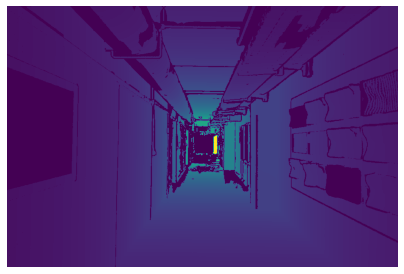} &
        \includegraphics[width=0.16\textwidth]{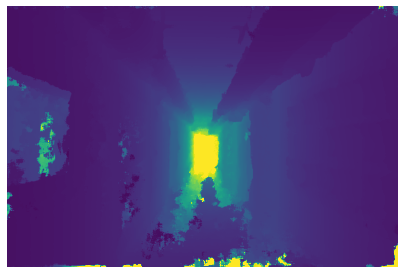} &
        \includegraphics[width=0.16\textwidth]{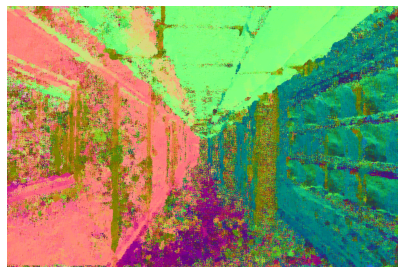} &
        \includegraphics[width=0.16\textwidth]{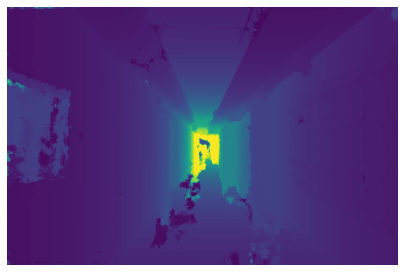} &
        \includegraphics[width=0.16\textwidth]{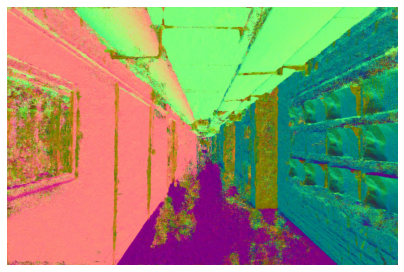}
        \\
        Ref. Image & GT. Depth & \multicolumn{2}{c}{PatchMatch-RL~\cite{lee2021patchmatchrl}} & \multicolumn{2}{c}{$\text{Ours}_{hr}$}
    \end{tabular}
    \caption{
        \textbf{Qualitative comparison against PatchMatch-RL on the ETH3D high-res multi-view benchmark training scenes.} 
        We show from the left the reference image, the ground-truth depth, the depth and normal estimated by PatchMatch-RL, the depth and normal estimated by $\text{Ours}_{hr}$. 
        All depth maps share the same color scale based on the ground-truth depth ranges. 
        Our method can reconstruct more accurate points in thin structure regions, and have less noise in texture-less surfaces.
    }
    \label{fig:eth3d} 
\end{figure*}

%% file: tables/eth3d.tex
\begin{table*}[t]
\centering
\caption{
\textbf{Results on the ETH3D high-res Multi-View benchmark train and test sets.} 
All methods do not train on any ETH3D data. 
Bold and underline denote the method with the highest and the second highest score for each setting, respectively. 
$\text{Ours}_{hr}$ method outperforms all other methods in both 2cm and 5cm criteria in train and test scenes except DeepC-MVS~\cite{kuhn2020deepc}, which was trained using the ETH3D train set. $\text{Ours}_{lr}$ outperforms all existing learning-based methods at all settings except for EPP-MVSNet~\cite{ma2021eppmvsnet} at the 2cm benchmark on the test set. 
} 
\resizebox{\textwidth}{!}{
\begin{tabular}{lcccccc@{\hskip 0.10in}ccccc@{\hskip 0.10in}ccccc@{\hskip 0.10in}ccccc@{\hskip 0.10in}ccccc@{\hskip 0.10in}ccccc}
\toprule
 & & \multicolumn{15}{c}{\textbf{Test 2cm: Accuracy / Completeness / F1}} & \multicolumn{15}{c}{\textbf{Test 5cm: Accuracy / Completeness / F1}} \\
\textbf{Method} & \textbf{Resolution} & \multicolumn{5}{c}{\textbf{Indoor}} & \multicolumn{5}{c}{\textbf{Outdoor}} & \multicolumn{5}{c}{\textbf{Combined}} & \multicolumn{5}{c}{\textbf{Indoor}} & \multicolumn{5}{c}{\textbf{Outdoor}} & \multicolumn{5}{c}{\textbf{Combined}} \\
\midrule
DeepC-MVS~\cite{kuhn2020deepc}                      & -  &         \multicolumn{5}{c}{89.1 / \S{85.2} / \F{86.9}}    & \multicolumn{5}{c}{89.4      / \S{86.4}    / \F{87.7}} & \multicolumn{5}{c}{89.2      / \S{85.5}    / \F{87.1}} & \multicolumn{5}{c}{95.3 / 91.1 / \S{93.0}}    & \multicolumn{5}{c}{95.7      / \S{92.9}    / \F{94.3}}   & \multicolumn{5}{c}{ 95.4      / 91.5        / \S{93.3}}\\
ACMP~\cite{Xu2020ACMP}                              & 3200x2130 & \multicolumn{5}{c}{90.6 / 74.2 / 80.6}        & \multicolumn{5}{c}{\S{90.4} / 79.6 / 84.4}        & \multicolumn{5}{c}{90.5 / 75.6 / 81.5}        & \multicolumn{5}{c}{95.4 / 83.2 / 88.4}        & \multicolumn{5}{c}{96.5 / 86.3 / 91.0}        & \multicolumn{5}{c}{95.7 / 84.0 / 89.0}    \\
ACMM~\cite{xu2019multi}                             & 3200x2130 & \multicolumn{5}{c}{91.0 / 72.7 / 79.8}        & \multicolumn{5}{c}{89.6 / 79.2 / 83.6}        & \multicolumn{5}{c}{\S{90.7} / 74.3 / 80.8}        & \multicolumn{5}{c}{96.1 / 82.9 / 88.5}        & \multicolumn{5}{c}{\S{97.0} / 86.3 / 91.1}        & \multicolumn{5}{c}{96.3 / 83.7 / 89.1}    \\
ACMH~\cite{xu2019multi}                             & 3200x2130 & \multicolumn{5}{c}{\S{91.1} / 64.8 / 73.9}        & \multicolumn{5}{c}{84.0 / 80.0 / 81.8}        & \multicolumn{5}{c}{89.3 / 68.6 / 75.9}        & \multicolumn{5}{c}{\F{97.4} / 75.0 / 83.7}        & \multicolumn{5}{c}{94.1 / 87.1 / 90.4}        & \multicolumn{5}{c}{\S{96.6} / 78.0 / 85.4}    \\
COLMAP~\cite{schoenberger2016mvs}                   & 3200x2130 & \multicolumn{5}{c}{\F{92.0} / 59.7 / 70.4}        & \multicolumn{5}{c}{\F{92.0} / 73.0 / 80.8}        & \multicolumn{5}{c}{\F{92.0} / 63.0 / 73.0}        & \multicolumn{5}{c}{\S{96.6} / 73.0 / 82.0}        & \multicolumn{5}{c}{\F{97.1} / 83.9 / 89.7}        & \multicolumn{5}{c}{\F{96.8} / 75.7 / 84.0}    \\
\midrule PatchmatchNet~\cite{wang2020patchmatchnet} & 2688x1792 & \multicolumn{5}{c}{68.8 / 74.6 / 71.3}        & \multicolumn{5}{c}{72.3 / 86.0 / 78.5}        & \multicolumn{5}{c}{69.7 / 77.5 / 73.1}        & \multicolumn{5}{c}{84.6 / 85.1 / 84.7}        & \multicolumn{5}{c}{87.0 / 92.0 / 89.3}        & \multicolumn{5}{c}{85.2 / 86.8 / 85.9}    \\
EPP-MVSNet~\cite{ma2021eppmvsnet}                   & 3072x2048 & \multicolumn{5}{c}{85.0 / 81.5 / 83.0}        & \multicolumn{5}{c}{86.8 / 82.6 / 84.6}        & \multicolumn{5}{c}{85.5 / 81.8 / 83.4}        & \multicolumn{5}{c}{93.7 / 89.7 / 91.6}        & \multicolumn{5}{c}{94.6 / 89.9 / 92.1}        & \multicolumn{5}{c}{93.9 / 89.8 / 91.7}    \\
PatchMatch-RL~\cite{lee2021patchmatchrl}            & 1920x1280 & \multicolumn{5}{c}{73.2 / 70.0 / 70.9}        & \multicolumn{5}{c}{78.3 / 78.3 / 76.8}        & \multicolumn{5}{c}{74.5 / 72.1 / 72.4}        & \multicolumn{5}{c}{88.0 / 83.7 / 85.5}        & \multicolumn{5}{c}{92.6 / 89.0 / 90.5}        & \multicolumn{5}{c}{89.2 / 85.0 / 86.8}    \\
$\text{Ours}_{lr}$                                  & 1920x1280 & \multicolumn{5}{c}{83.8 / 81.5 / 82.1}        & \multicolumn{5}{c}{85.8 / 83.5 / 84.1}        & \multicolumn{5}{c}{84.3 / 82.0 / 82.6}        & \multicolumn{5}{c}{93.3 / \S{91.6} / 92.3}        & \multicolumn{5}{c}{94.8 / 91.5 / 93.0}        & \multicolumn{5}{c}{93.7 / \S{91.6} / 92.5}    \\
$\text{Ours}_{hr}$                                  & 3200x2112 & \multicolumn{5}{c}{84.3 / \F{85.8} / \S{84.8}}    & \multicolumn{5}{c}{85.5 / \F{87.4} / \S{86.3}}    & \multicolumn{5}{c}{84.6 / \F{86.2} / \S{85.1}}    & \multicolumn{5}{c}{93.4 / \F{93.4}  / \F{93.3}}    & \multicolumn{5}{c}{94.3 / \F{93.5} / \S{93.8}}    & \multicolumn{5}{c}{93.6 / \F{93.4} / \F{93.5}}\\
\midrule
\\
\midrule
 &  & \multicolumn{15}{c}{\textbf{Train 2cm: Accuracy / Completeness / F1}} & \multicolumn{15}{c}{\textbf{Train 5cm: Accuracy / Completeness / F1}} \\
\textbf{Method} & \textbf{Resolution} & \multicolumn{5}{c}{\textbf{Indoor}} & \multicolumn{5}{c}{\textbf{Outdoor}} & \multicolumn{5}{c}{\textbf{Combined}} & \multicolumn{5}{c}{\textbf{Indoor}} & \multicolumn{5}{c}{\textbf{Outdoor}} & \multicolumn{5}{c}{\textbf{Combined}} \\
\midrule
ACMP~\cite{Xu2020ACMP}                              & 3200x2130 & \multicolumn{5}{c}{92.3 / 72.3 / \S{80.5}}    & \multicolumn{5}{c}{87.6 / 72.0 / \S{78.9}}    & \multicolumn{5}{c}{90.1 / 72.2 / \S{79.8}}    & \multicolumn{5}{c}{96.3 / 81.8 / 88.1}        & \multicolumn{5}{c}{95.6 / 82.7 / 88.6}        & \multicolumn{5}{c}{96.0 / 82.2 / 88.3}    \\
ACMM~\cite{xu2019multi}                             & 3200x2130 & \multicolumn{5}{c}{92.5 / 68.5 / 78.1}        & \multicolumn{5}{c}{\F{88.6} / \S{72.7} / \F{79.7}}    & \multicolumn{5}{c}{\S{90.7} / 70.4 / 78.9}        & \multicolumn{5}{c}{96.4 / 78.4 / 86.1}        & \multicolumn{5}{c}{\F{96.2} / 83.9 / \S{89.5}}    & \multicolumn{5}{c}{96.3 / 80.9 / 87.7}    \\
ACMH~\cite{xu2019multi}                             & 3200x2130 & \multicolumn{5}{c}{\S{92.6} / 59.2 / 70.0}        & \multicolumn{5}{c}{84.7 / 64.4 / 71.5}        & \multicolumn{5}{c}{88.9 / 61.6 / 70.7}        & \multicolumn{5}{c}{\S{97.7} / 70.1 / 80.5}        & \multicolumn{5}{c}{95.4 / 75.6 / 83.5}        & \multicolumn{5}{c}{\S{96.6} / 72.7 / 81.9}    \\
COLMAP~\cite{schoenberger2016mvs}                   & 3200x2130 & \multicolumn{5}{c}{\F{95.0} / 52.9 / 66.8}        & \multicolumn{5}{c}{\S{88.2} / 57.7 / 68.7}        & \multicolumn{5}{c}{\F{91.9} / 55.1 / 67.7}        & \multicolumn{5}{c}{\F{98.0} / 66.6 / 78.5}        & \multicolumn{5}{c}{\S{96.1} / 73.8 / 82.9}        & \multicolumn{5}{c}{\F{97.1} / 69.9 / 80.5}    \\
\midrule PatchmatchNet~\cite{wang2020patchmatchnet} & 2688x1792 & \multicolumn{5}{c}{63.7 / 67.7 / 64.7}        & \multicolumn{5}{c}{66.1 / 62.8 / 63.7}        & \multicolumn{5}{c}{64.8 / 65.4 / 64.2}        & \multicolumn{5}{c}{78.7 / 80.0 / 78.9}        & \multicolumn{5}{c}{86.8 / 73.2 / 78.5}        & \multicolumn{5}{c}{82.4 / 76.9 / 78.7}    \\
EPP-MVSNet~\cite{ma2021eppmvsnet}                   & 3072x2048 & \multicolumn{5}{c}{86.5 / 71.1 / 77.4}        & \multicolumn{5}{c}{78.5 / 63.5 / 70.0}        & \multicolumn{5}{c}{82.8 / 67.6 / 74.0}        & \multicolumn{5}{c}{94.0 / 82.8 / 87.8}        & \multicolumn{5}{c}{93.3 / 78.5 / 85.2}        & \multicolumn{5}{c}{93.6 / 80.8 / 86.6}    \\
PatchMatch-RL~\cite{lee2021patchmatchrl}            & 1920x1280 & \multicolumn{5}{c}{76.6 / 60.7 / 66.7}        & \multicolumn{5}{c}{75.4 / 64.0 / 69.1}        & \multicolumn{5}{c}{76.1 / 62.2 / 67.8}        & \multicolumn{5}{c}{89.6 / 76.5 / 81.4}        & \multicolumn{5}{c}{88.8 / 81.4 / 85.7}        & \multicolumn{5}{c}{90.5 / 78.8 / 83.3}    \\
$\text{Ours}_{lr}$                                  & 1920x1280 & \multicolumn{5}{c}{85.3 / \S{76.0} / 79.9}        & \multicolumn{5}{c}{80.7 / 72.3 / 76.1}        & \multicolumn{5}{c}{83.2 / \S{74.3} / 78.1}        & \multicolumn{5}{c}{93.7 / \S{88.7} / \S{90.9}}    & \multicolumn{5}{c}{92.6 / \S{86.3} / 89.3}        & \multicolumn{5}{c}{93.2 / \S{87.6} / \S{90.1}}\\
$\text{Ours}_{hr}$                                  & 3200x2112 & \multicolumn{5}{c}{86.1 / \F{78.4} / \F{81.5}}    & \multicolumn{5}{c}{81.4 / \F{76.3} / 78.6}        & \multicolumn{5}{c}{84.0 / \F{77.4} / \F{80.1}}    & \multicolumn{5}{c}{94.3 / \F{88.8} / \F{91.3}}    & \multicolumn{5}{c}{93.3 / \F{86.6} / \F{89.7}}    & \multicolumn{5}{c}{93.8 / \F{87.8} / \F{90.6}}\\
\bottomrule
\end{tabular}
}
\label{tab:eth3d}
\end{table*}

%% file: figures/pointcloud_results.tex
\begin{figure*}[t]
    \centering
    \begin{tabular}{ccc}
    \includegraphics[width=0.31\textwidth]{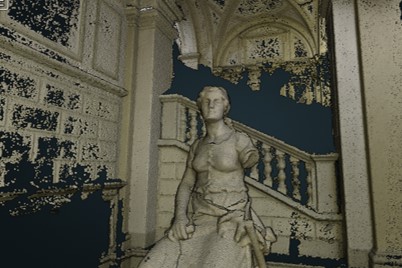} &
    \includegraphics[width=0.31\textwidth]{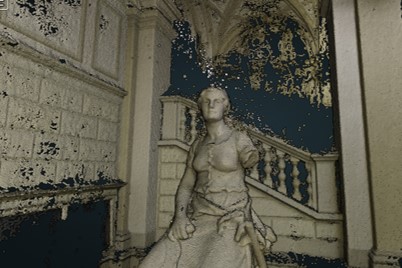} &
    \includegraphics[width=0.31\textwidth]{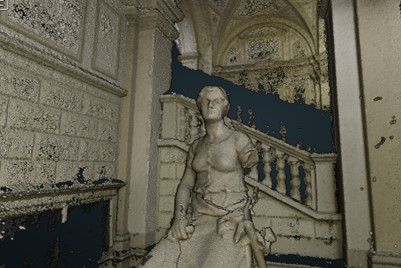} \\
    ACMM~\cite{xu2019multi} & EPP-MVS~\cite{ma2021eppmvsnet} & Ours
    \end{tabular}
    \caption{
    \textbf{Point cloud reconstruction results on ETH3D Dataset.}
    From left to right, we show the reconstruction results of ACMM~\cite{xu2019multi}, EPP-MVSNet~\cite{ma2021eppmvsnet} and our method in the \textit{Statue} from the ETH3D benchmark~\cite{schoeps2017eth3d}. We refer to the \href{https://www.eth3d.net/result_details?id=302}{benchmark website} for all reconstruction results on the benchmark scenes.
    }
    \label{fig:eth_epp_acmm} 
\end{figure*}

%% file: tables/tanks_and_temples.tex
\begin{table*}[t]
\centering
\caption{
    \textbf{Results on the Tanks and Temples benchmark.} 
    We show precision, recall and $F_1$ scores on the intermediate and advanced scenes for each methods. 
    The best performing model is marked as bold. 
}
\begin{tabular}{lccc}
\toprule
& \multicolumn{3}{c}{\textbf{Precision / Recall / F1}}\\
\multicolumn{1}{c}{\textbf{Method}} & 
\multicolumn{1}{c}{\textbf{Intermediate}} && \multicolumn{1}{c}{\textbf{Advanced}} \\
\midrule
DeepC-MVS~\cite{kuhn2020deepc}             & 59.1 / 61.2 / 59.8 && 40.7 / 31.3 / 34.5\\
ACMP~\cite{Xu2020ACMP}                     & 49.1 / 73.6 / 58.4 && \textBF{42.5} / 34.6 / \textBF{37.4} \\
ACMM~\cite{xu2019multi}                    & 49.2 / 70.9 / 57.3 && 35.6 / 34.9 / 34.0 \\
COLMAP~\cite{schoenberger2016mvs}          & 43.2 / 44.5 / 42.1 && 33.7 / 24.0 / 27.2 \\
R-MVSNet~\cite{yao2019recurrent}           & 43.7 / 57.6 / 48.4 && 31.5 / 22.1 / 24.9 \\
CasMVSNet~\cite{gu2020cascade}             & 47.6 / 74.0 / 56.8 && 29.7 / 35.2 / 31.1 \\
AttMVS~\cite{luo2020attention}             & \textBF{61.9} / 58.9 / 60.1 && 40.6 / 27.3 / 31.9 \\
PatchmatchNet~\cite{wang2020patchmatchnet} & 43.6 / 69.4 / 53.2 && 27.3 / \textBF{41.7} / 32.3 \\
EPP-MVSNet~\cite{ma2021eppmvsnet}          & 53.1 / \textBF{75.6} / \textBF{61.7} && 34.6 / 35.6 / 35.7 \\
PatchMatch-RL~\cite{lee2021patchmatchrl}   & 45.9 / 62.3 / 51.8 && 30.6 / 36.8 / 31.8 \\
$\text{Ours}$                              & 50.5 / 63.3 / 55.3 && 38.8 / 32.4 / 33.8 \\
\bottomrule
\end{tabular}
\label{tab:tnt}
\end{table*}

%% file: figures/tanks_and_temples.tex
\begin{figure*}[t]
    \centering
        \begin{tabular}{cc}
        \includegraphics[trim={0 8em 0 7em},clip,width=0.48\textwidth]{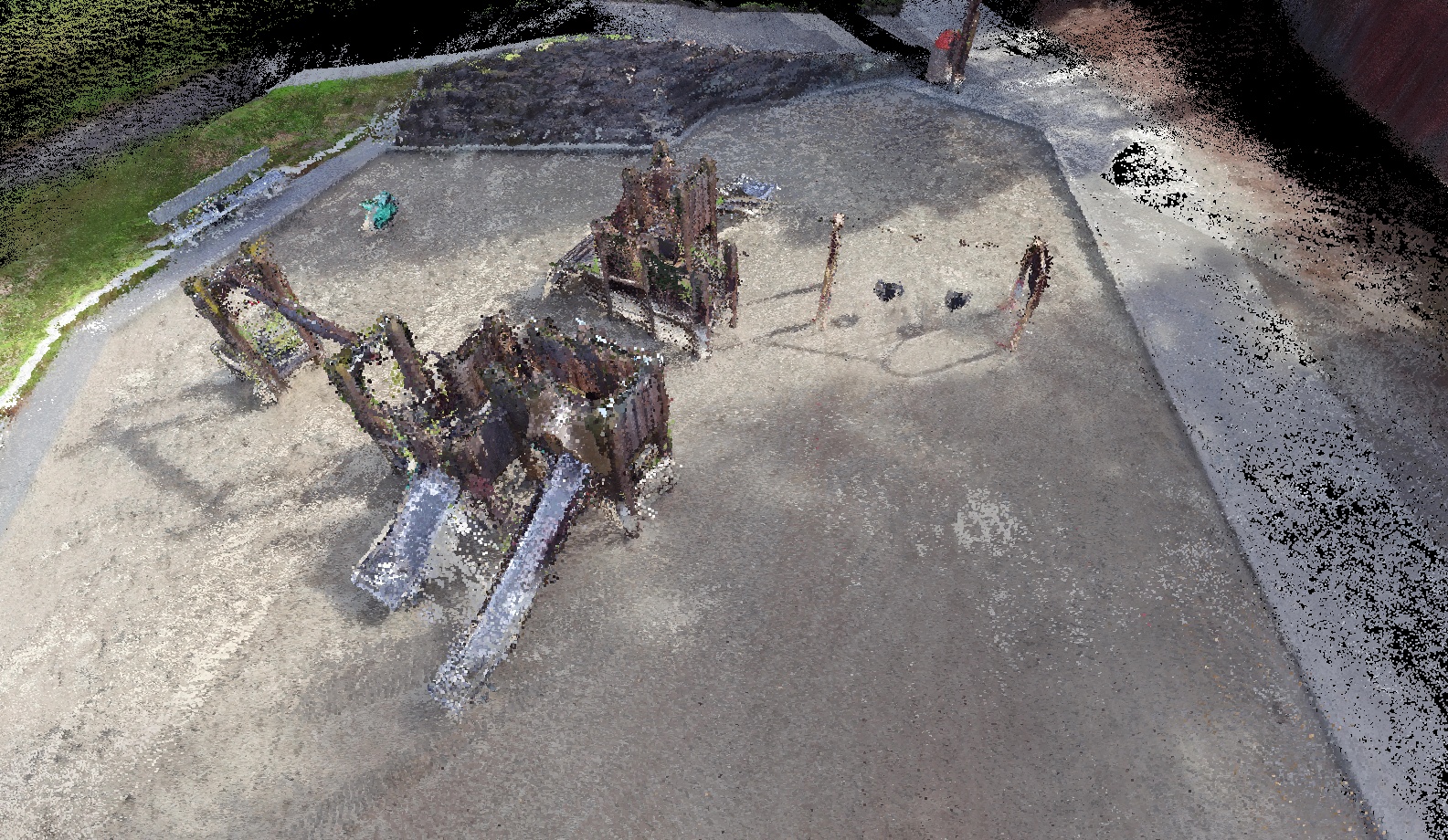} & 
        \includegraphics[trim={0 8em 0 7em},clip,width=0.48\textwidth]{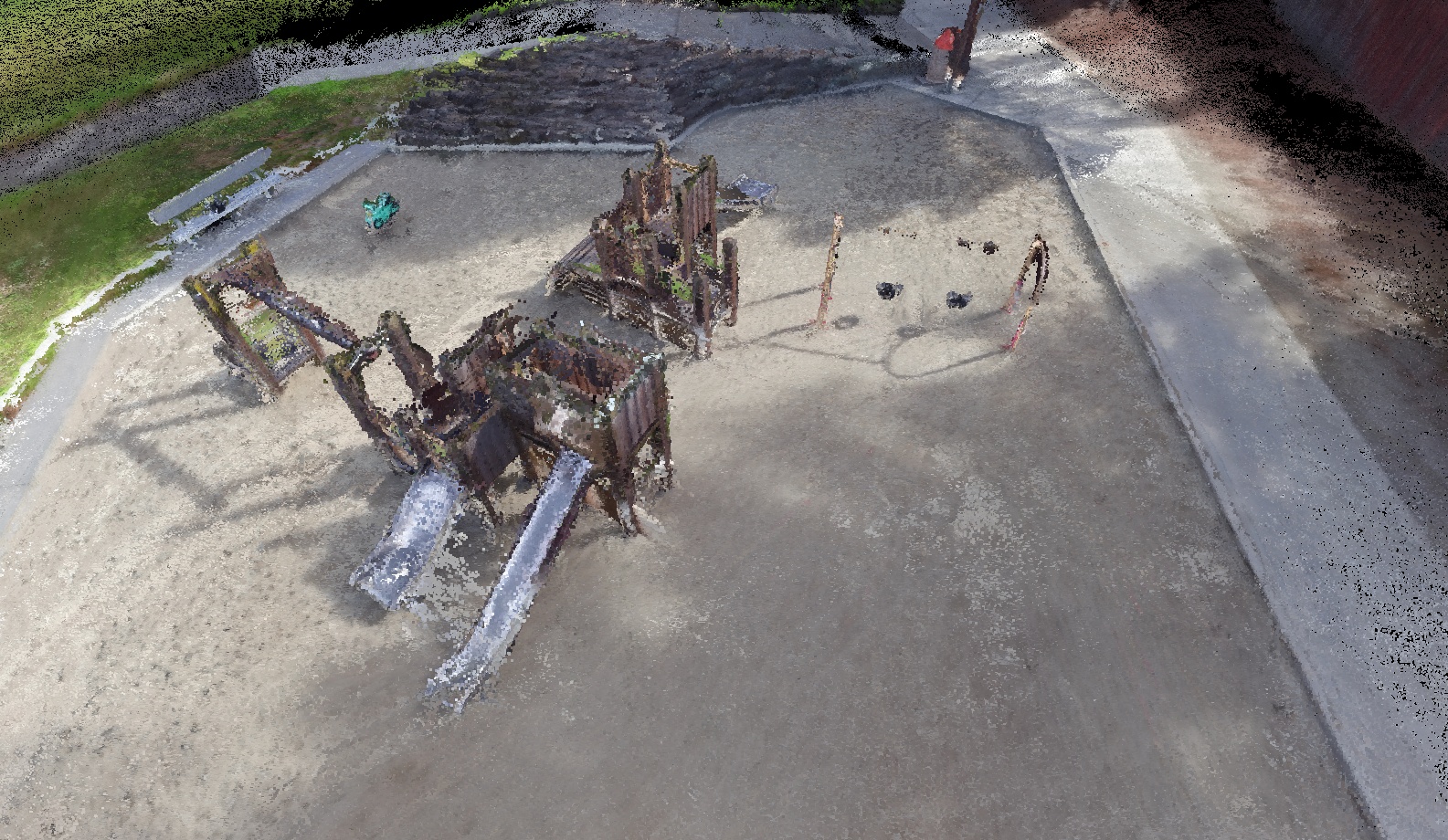}\\
        \includegraphics[trim={0 5em 0  10em},clip,width=0.48\textwidth]{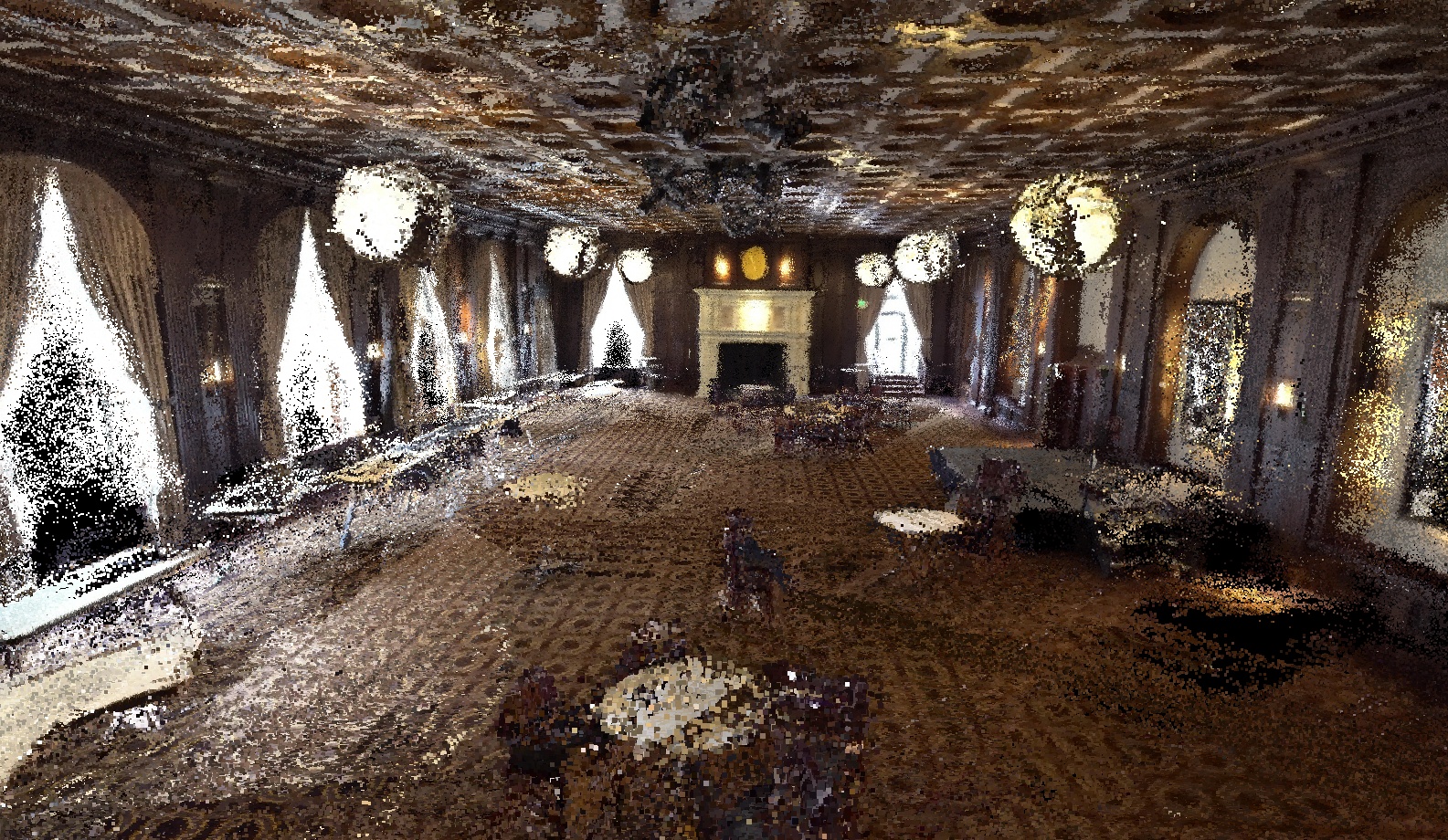} &
        \includegraphics[trim={0 5em 0 10em},clip,width=0.48\textwidth]{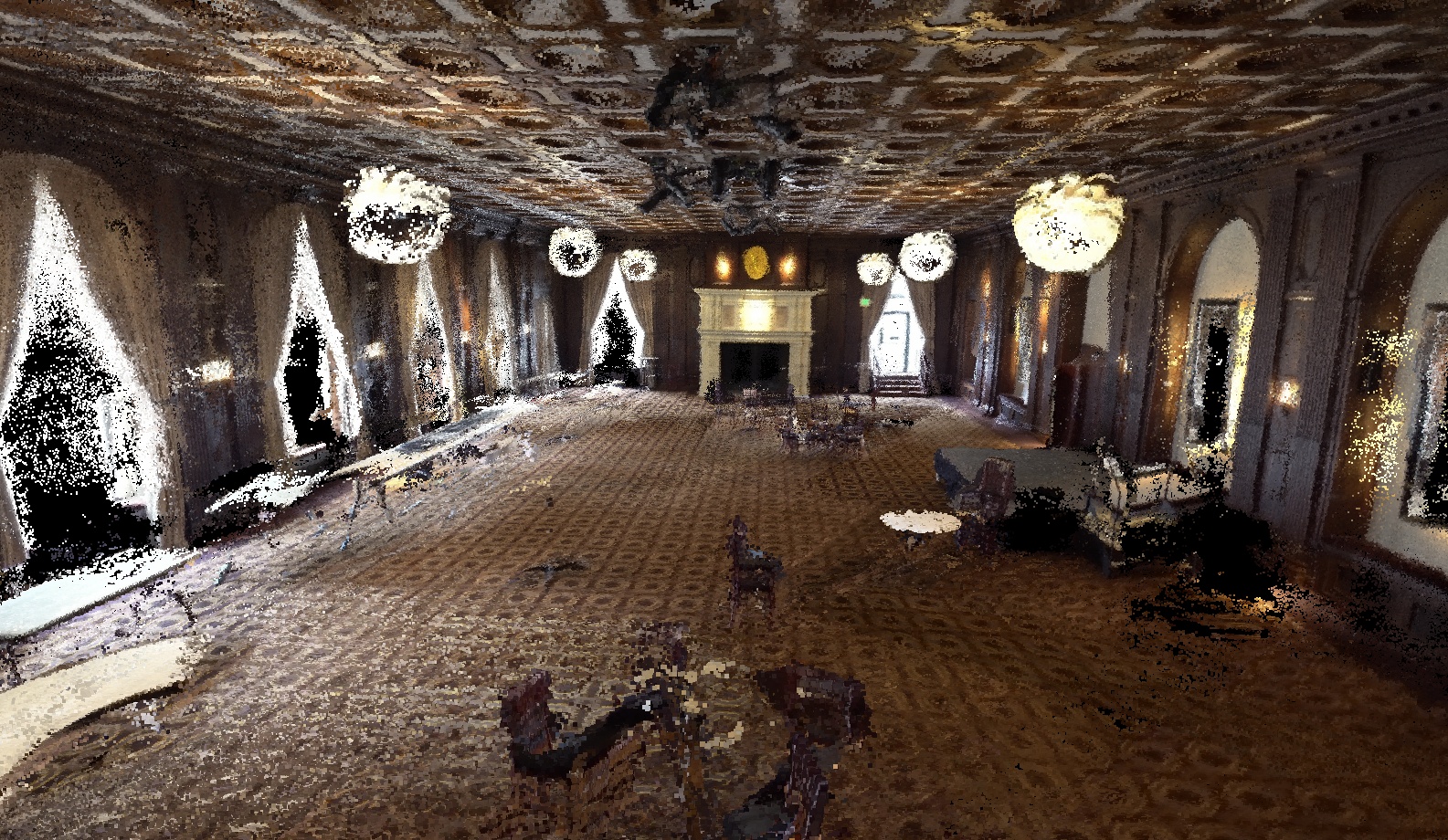}\\
        PatchMatch-RL~\cite{lee2021patchmatchrl} & Ours
    \end{tabular}
    \caption{
    \textbf{Point Cloud reconstruction results on Tanks and Temples Dataset}
    We show PatchMatch-RL~\cite{lee2021patchmatchrl} and Our reconstruction results on the \textit{Playground} and \textit{Ballroom}  scene from Tanks and Temples dataset. 
    Our method achieves more detail and less noise.
    }
    \label{fig:tnt_qual} 
\end{figure*}

%% file: tables/ablations.tex
\begin{table*}[t]
\centering
\caption{
    \textbf{Ablation Study on ETH3D High-Res Training Set at lower resolution 1920 × 1280 ($\text{Ours}_{lr}$).} 
    We compare how much each module contributes to the overall reconstruction quality. 
    ``G. C.'' denotes geometric consistency, ``U-Net'' denotes the feature extractor using shallow VGG-style feature extractor, and ``Copl.'' denotes dot-product attention instead of learned coplanarity to weight supporting pixels. The memory measures peak GPU reserved memory, and time measures average time used for inference on one reference image. 
    We use the author provided values for PatchMatch-RL~\cite{lee2021patchmatchrl}.
    The model with the highest score is marked with bold. 
}
\resizebox{\textwidth}{!}{
\begin{tabular}{cccccccc}
    \toprule
    \multicolumn{3}{c}{\textbf{Model}}  & \multicolumn{3}{c}{\textbf{Accuracy / Completeness / F1}} & \multirow{ 2}{*}{\textbf{Mem. (MiB)}}& \multirow{2}{*}{\textbf{Time (s)}}     \\
    \textbf{G. C.} & \textbf{U-Net} & \textbf{Copl.} & \textbf{Train 2CM} &\phantom{a}& \textbf{Train 5cm} & & \\
    \midrule
    \multicolumn{3}{c}{PatchMatch-RL}          & 76.1 / 62.2 / 67.8                   && 90.5 / 78.8 / 83.3                    & 7,693 & 13.54 \\
$\times$     & $\times$     & $\times$     & \textBF{84.1} / 64.7 / 72.3          && \textBF{94.0} / 78.8 / 85.2           & \textbf{4,520}          & \textbf{5.02}         \\
$\checkmark$ & $\times$     & $\times$     & 83.9 / 67.0 / 73.8                   && 93.6 / 80.7 / 86.3                    & 4,952          & 8.78         \\
$\times$     & $\checkmark$ & $\times$     & 84.0 / 70.5 / 76.1                   && 93.9 / 84.6 / 88.8                    & 4,566          & 5.43         \\
$\times$     & $\times$     & $\checkmark$ & 82.8 / 67.2 / 73.5                   && 92.9 / 81.0 / 86.2                    & 5,058          & 6.68         \\
$\times$     & $\checkmark$ & $\checkmark$ & 83.6 / 73.3 / 77.3                   && 93.3 / 86.8 / 89.8                    & 5,092          & 7.00          \\
$\checkmark$ & $\times$     & $\checkmark$ & 82.8 / 72.9 / 77.0                   && 92.8 / 86.2 / 89.2                    & 5,320          & 10.61         \\
$\checkmark$ & $\checkmark$ & $\times$     & 83.3 / 73.2 / 77.2                   && 93.2 / 86.7 / 89.6                    & 5,012          & 9.22          \\
$\checkmark$ & $\checkmark$ & $\checkmark$ & 83.2 / \textBF{74.3} / \textBF{78.1} && 93.2 / \textBF{87.6} / \textBF{90.1}  & 5,658         & 10.98         \\
    \bottomrule
\end{tabular}
}
\label{tab:ablation}
\end{table*}

%% file: tables/supp/ablation_hr.tex
\begin{table*}[h!]
\centering
\caption{
\textbf{Ablation Study on ETH3D High-Res Training Set at higher resolution 3200 × 2112 ($\text{Ours}_{hr}$).} 
We let ``Geo. Cons.'' to denote the geometric consistency, ``U-Net'' to denote the feature extractor using shallow VGG-style feature extractor, and ``Copl.'' to denote the method using dot-product attention instead of learned coplanarity to weight supporting pixels. 
}
\label{tab:ablation_hr}
\resizebox{\textwidth}{!}{
\begin{tabular}{cccccccc}
    \toprule
    \multicolumn{3}{c}{\textbf{Model}}  & \multicolumn{3}{c}{\textbf{Accuracy / Completeness / F1}} & \multirow{ 2}{*}{\textbf{Mem. (MiB)}}& \multirow{2}{*}{\textbf{Time (s)}}     \\
    \textbf{G. C.} & \textbf{U-Net} & \textbf{Copl.} & \textbf{Train 2CM} &\phantom{a}& \textbf{Train 5cm} & & \\
\midrule
$\times$        & $\times$      & $\times$      & \textbf{85.4} / 68.2  / 75.8          && \textbf{94.8} / 80.7 / 87.1          &  \textbf{8,502} & \textbf{14.03} \\
$\checkmark$    & $\times$      & $\times$      & 84.8 / 70.8 / 77.4          && 94.4 / 83.4 / 88.3          & 12,160 & 22.12 \\
$\times$        & $\checkmark$  & $\times$      & 85.2 / 73.1 / 78.1          && 93.8 / 84.8 / 88.9          &  8,514 & 16.52 \\
$\times$        & $\times$      & $\checkmark$  & 83.6 / 70.5 / 77.2          && 93.9 / 83.5 / 88.1          & 10,050 & 19.13 \\
$\times$        & $\checkmark$  & $\checkmark$  & 84.3 / 76.0 / 79.4          && 94.1 / 86.8 / 90.1          &  9,523 & 20.25 \\
$\checkmark$    & $\times$      & $\checkmark$  & 83.5 / 75.7 / 78.7          && 93.3 / 86.1 / 89.3          & 12,982 & 27.71 \\
$\checkmark$    & $\checkmark$  & $\times$      & 83.9 / 76.1 / 79.1          && 93.9 / 86.2 / 89.6          & 12,230 & 24.89 \\
$\checkmark$    & $\checkmark$  & $\checkmark$  & 84.0 / \textbf{77.4} / \textbf{80.1} && 93.8 / \textbf{87.8} / \textbf{90.6} & 13,484 & 28.65 \\
\bottomrule
\end{tabular}
}
\end{table*}

%% file: tables/supp/ablation_additional.tex
\begin{table*}[h!]
\centering
\caption{
\textbf{Additional Ablation Studies.} 
``Bilinear-JPEG'' denotes the image downsampled with bilinear sampling and stored with JPEG format; ``Trained with Less-View \& Iter.'', the model trained with the same hyperparameters as PatchMatch-RL~\cite{lee2021patchmatchrl}; 
``No Fusion Upsampling'', the model fused without nearest-neighbor upsampling followed by median filtering; ``Adapt-3'' and ``Adapt-5'', inference with 3 and 5 adaptively sampled supporting pixels; and
``Without CUDA Kernel``, inference without using the custom CUDA kernel. 
The model with the highest score is marked with bold.
}
\label{tab:ablation_additional}
\resizebox{\textwidth}{!}{
\begin{tabular}{lccccc@{\hskip 0.6in}ccccccc}
\toprule
& \multicolumn{10}{c}{\textbf{Accuracy / Completeness / F1}} & \multirow{ 2}{*}{\textbf{Mem. (MiB)}}& \multirow{2}{*}{\textbf{Time (s)}}     \\
\textbf{Model}    & \multicolumn{5}{c}{\textbf{Train 2CM}} & \multicolumn{5}{c}{\textbf{Train 5cm}} & & \\
\midrule
Bilinear-JPEG                  & 79.9 &/& 66.0 &/& 71.6                   & 91.3 &/& 81.7 &/& 85.9                     & \textbf{5,658} & 11.02 \\
Trained with Less-View \& Iter. & 81.9 &/& 72.2 &/& 76.3                   & 92.7 &/& 86.4 &/& 89.3                     & \textbf{5,658} & 11.02 \\
No Fusion Upsampling            & \textbf{85.6} &/& 66.8 &/& 74.8          & \textbf{94.2} &/& 85.6 &/& 89.6            & \textbf{5,658} & 10.92 \\
Adapt-3                        & 84.3 &/& 69.6 &/& 75.7                   & 94.1 &/& 84.7 &/& 89.1                     & \textbf{5,658}  & \textbf{9.45} \\
Adapt-5                        & 84.1 &/& 73.2 &/& 77.8                   & 93.9 &/& 87.0 &/& \textbf{90.1}            & \textbf{5,658} & 10.09 \\
Without CUDA Kernel             & 83.2 &/& \textbf{74.3} &/& \textbf{78.1} & 93.2 &/& \textbf{87.6} &/& \textbf{90.1}   & 10,839 & 24.33 \\
$\text{Ours}_{lr}$              & 83.2 &/& \textbf{74.3} &/& \textbf{78.1} & 93.2 &/& \textbf{87.6} &/& \textbf{90.1}   &  \textbf{5,658}  & 10.98 \\
\bottomrule
\end{tabular}
}
\end{table*}

%% file: figures/supp/jpeg_artifact.tex
\begin{figure*}[h]
    \centering
    \includegraphics[width=0.98\textwidth]{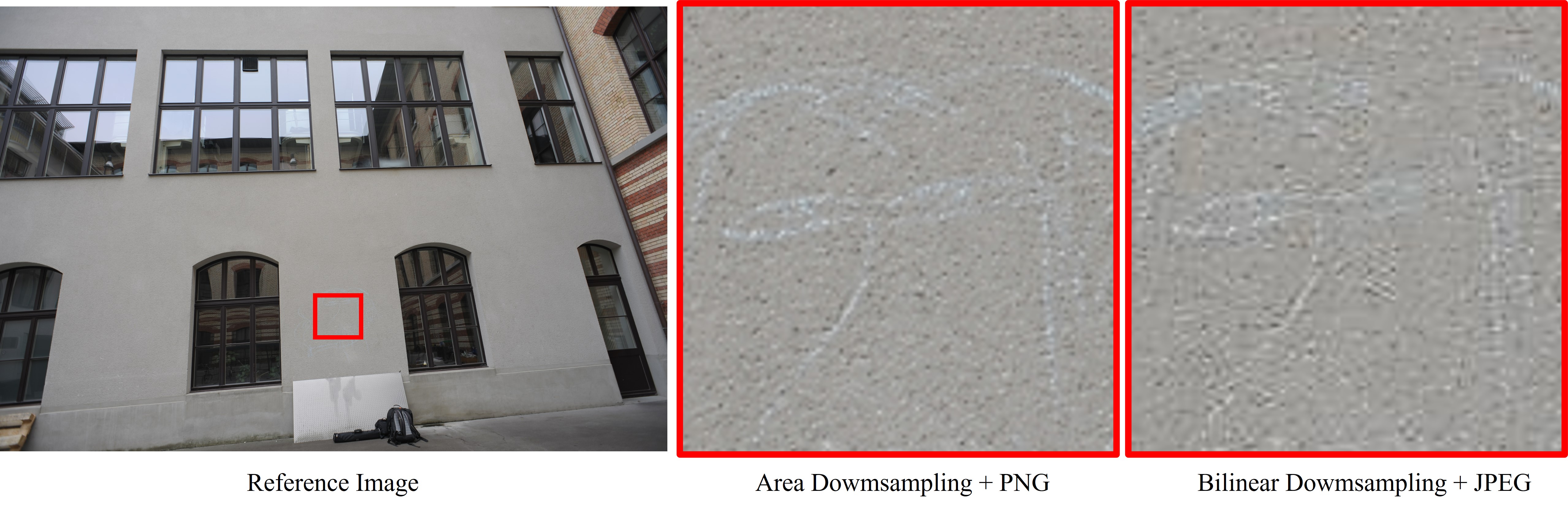}
    \caption{\textbf{Texture Artifact from JPEG compression.} The left image shows the reference image, and the right two shows image downsampled with area interpolation stored with PNG format and  image downsampled with bilinear interpolation stored with JPEG format. We show that there is an artifact in texture-light area in JPEG stored files, which causes huge drop in accuracy as shown in Table~\ref{tab:ablation_additional}.
    }
    \label{fig:downsample_artifact} 
\end{figure*}

%% file: figures/supp/eth3d_hr_train.tex
\begin{figure*}[ht]
    \centering
    \includegraphics[trim={0 5em 0 10em},clip,width=0.48\textwidth]{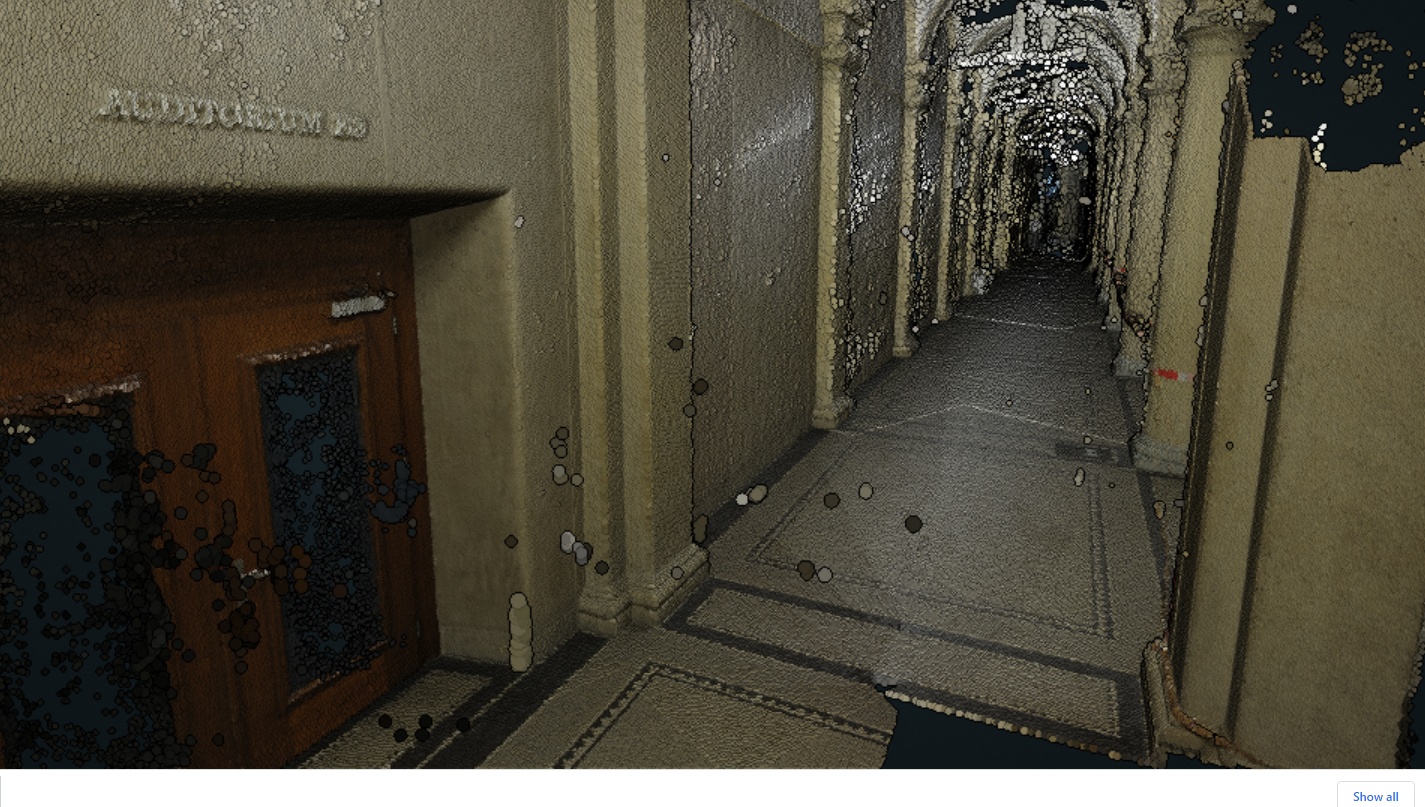}
    \includegraphics[trim={0 5em 0 10em},clip,width=0.48\textwidth]{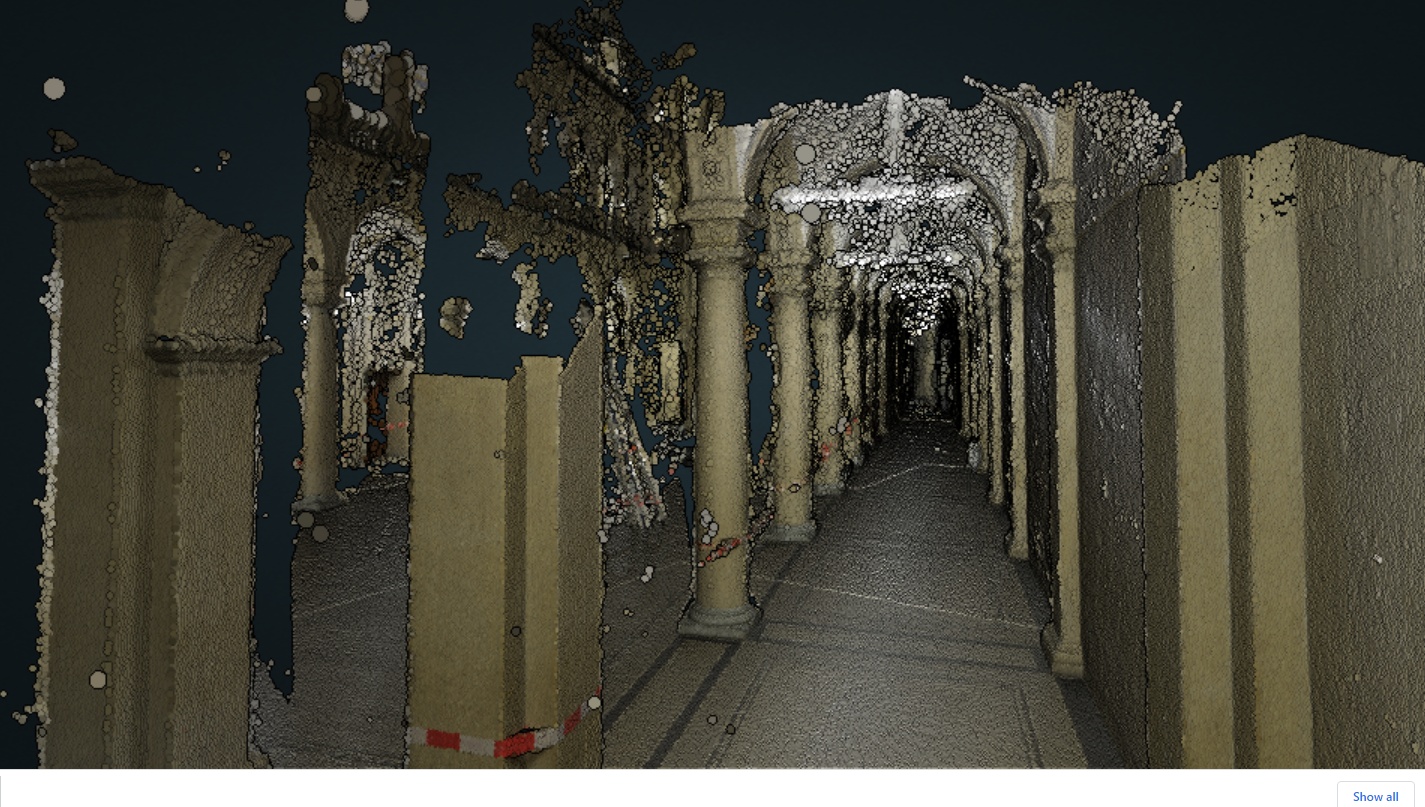}\\
    \includegraphics[trim={0 5em 0 10em},clip,width=0.48\textwidth]{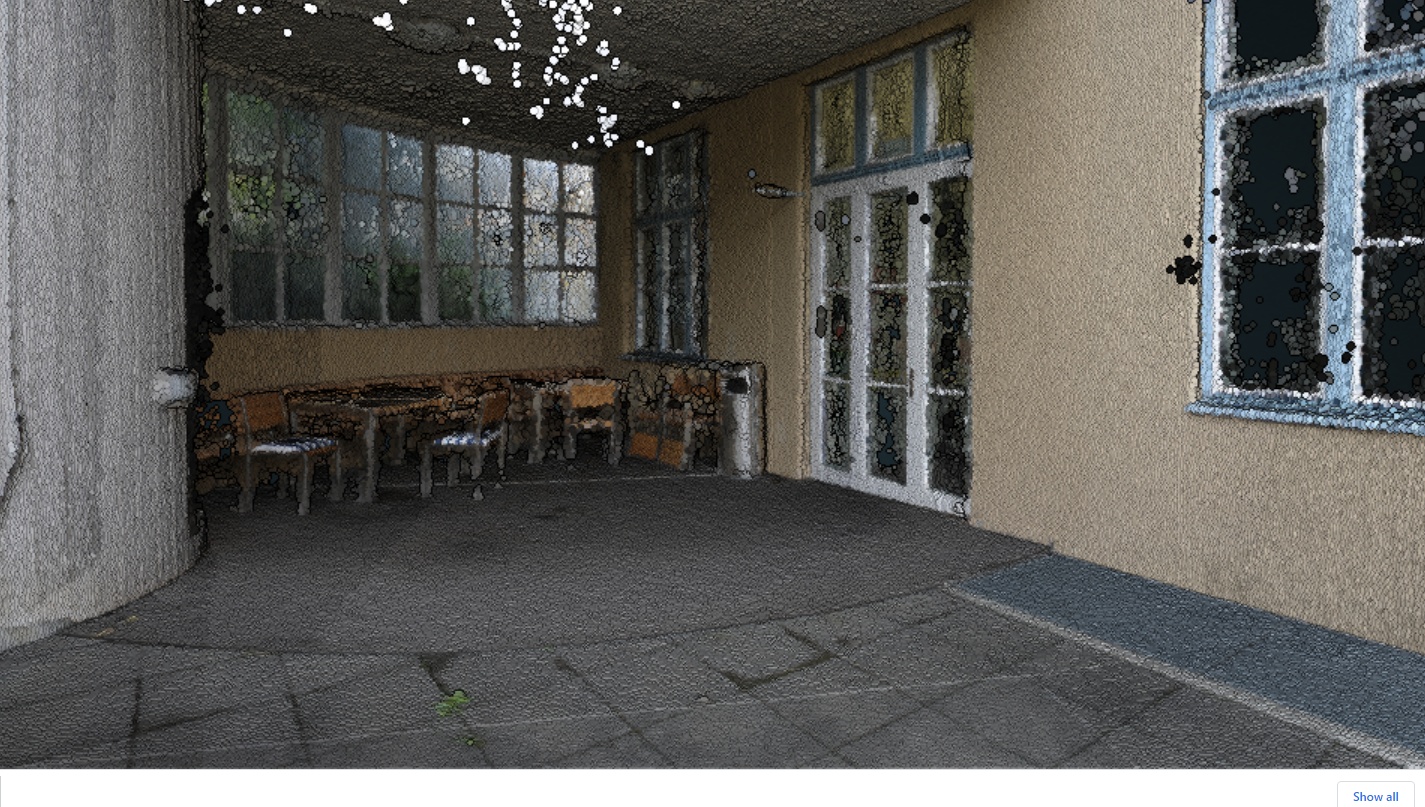}
    \includegraphics[trim={0 5em 0 10em},clip,width=0.48\textwidth]{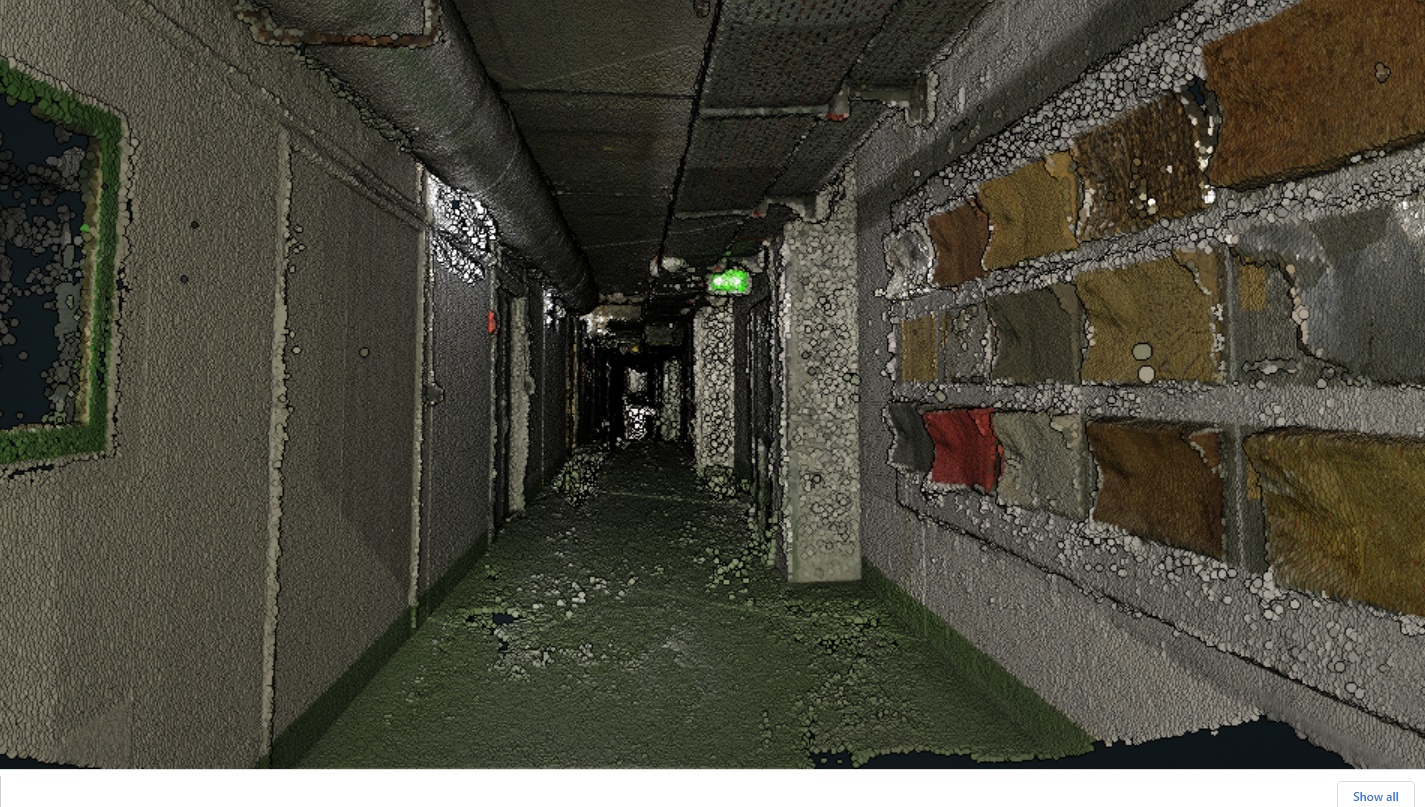}\\
    \includegraphics[trim={0 5em 0 10em},clip,width=0.48\textwidth]{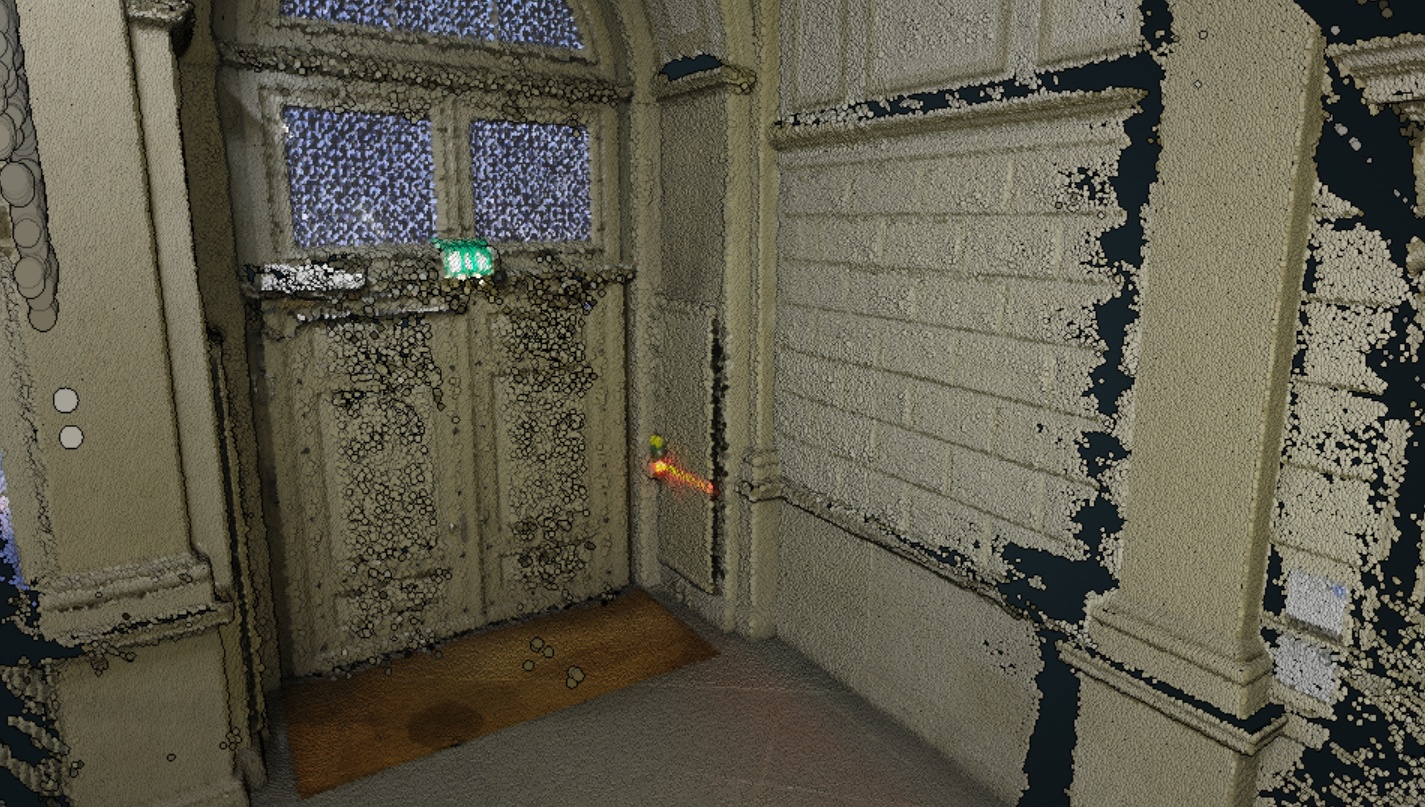}
    \includegraphics[trim={0 5em 0 10em},clip,width=0.48\textwidth]{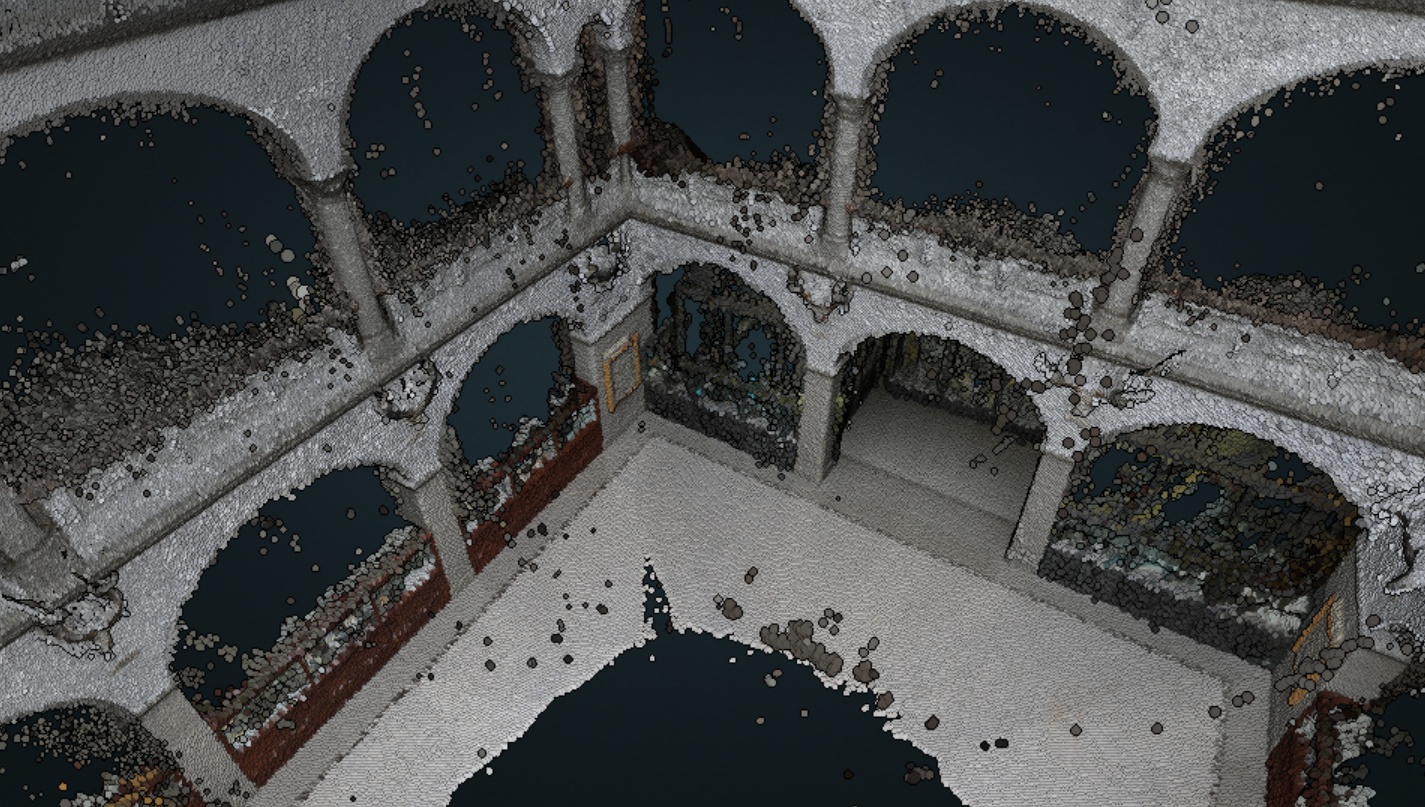}\\
    \includegraphics[trim={0 5em 0 10em},clip,width=0.48\textwidth]{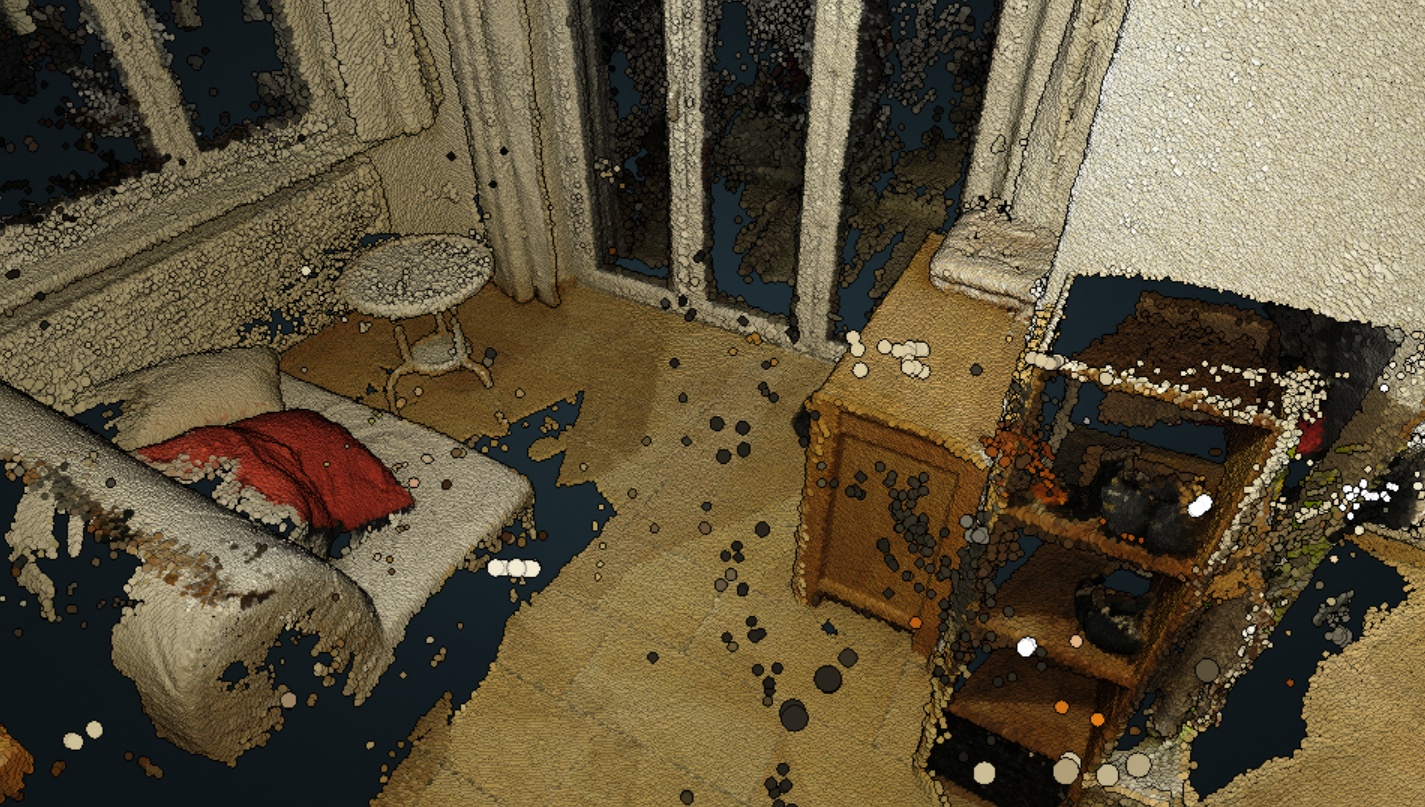}
    \includegraphics[trim={0 5em 0 10em},clip,width=0.48\textwidth]{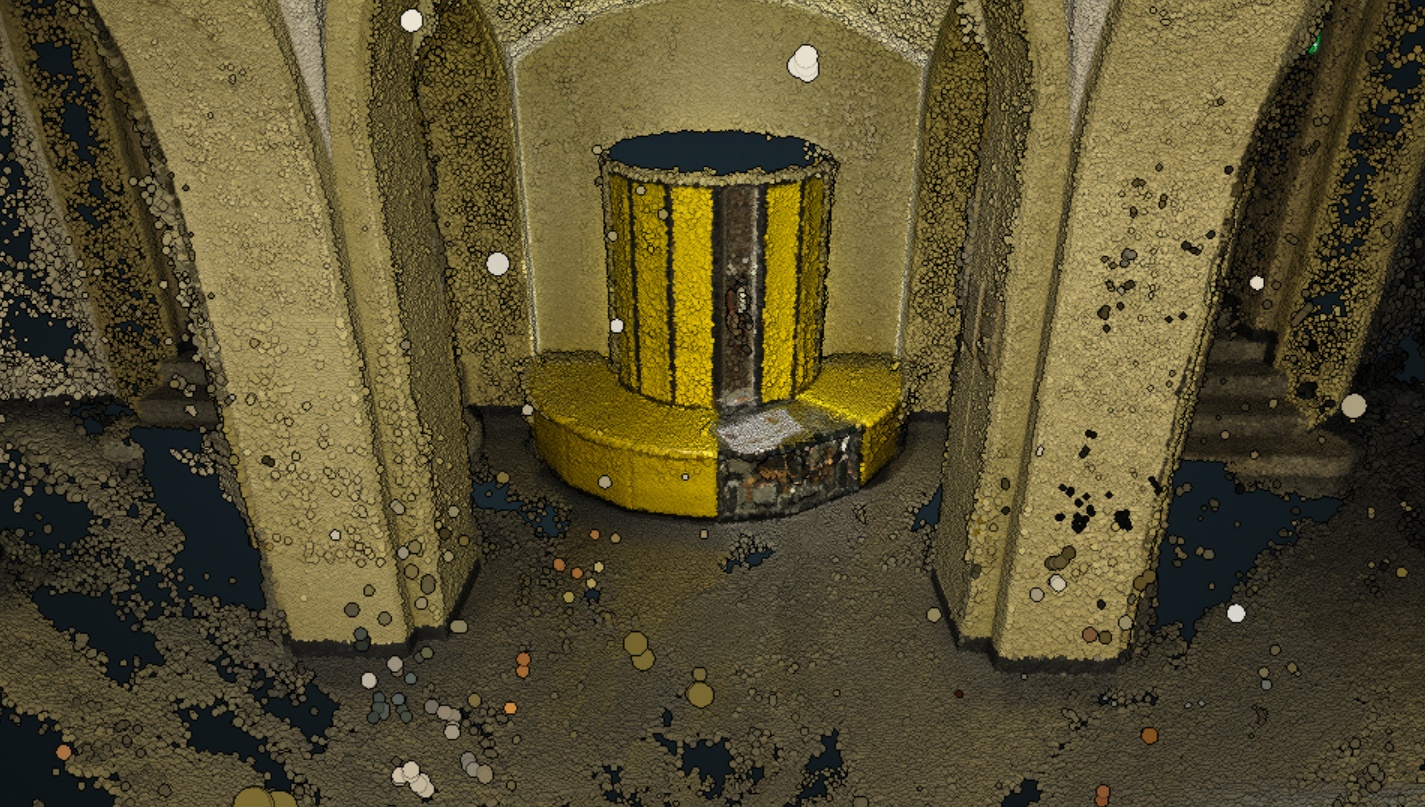}\\
    \caption{ \textbf{Additional Qualitative Results for ETH3D High-Res~\cite{schoeps2017eth3d} Train set (top two rows) and Test set (bottom two rows) for $\text{Ours}_{hr}$}.}
    \label{fig:eth_hr_train} 
\end{figure*}

%% file: figures/supp/eth3d_lr_train.tex
\begin{figure*}[ht]
    \centering
    \includegraphics[trim={0 5em 0 10em},clip,width=0.48\textwidth]{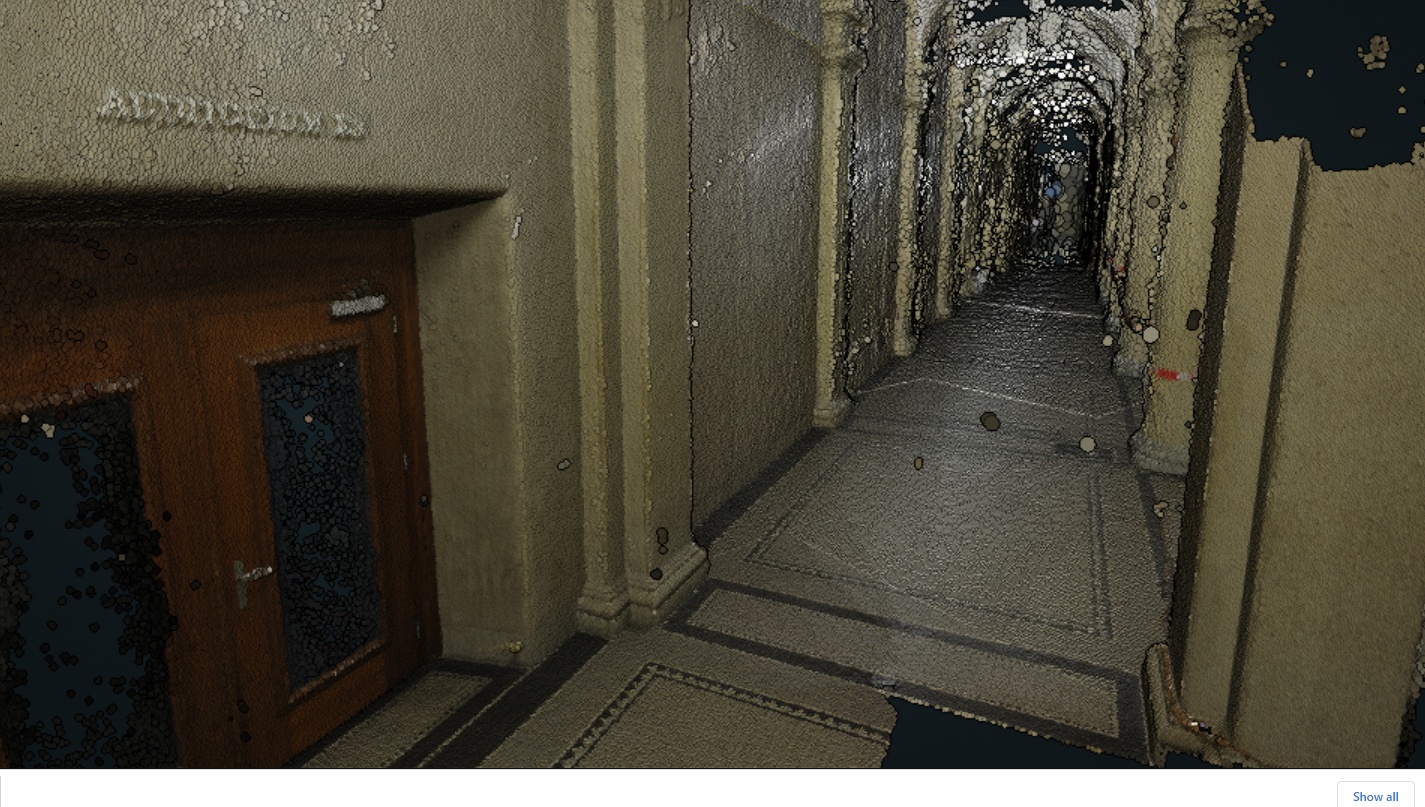}
    \includegraphics[trim={0 5em 0 10em},clip,width=0.48\textwidth]{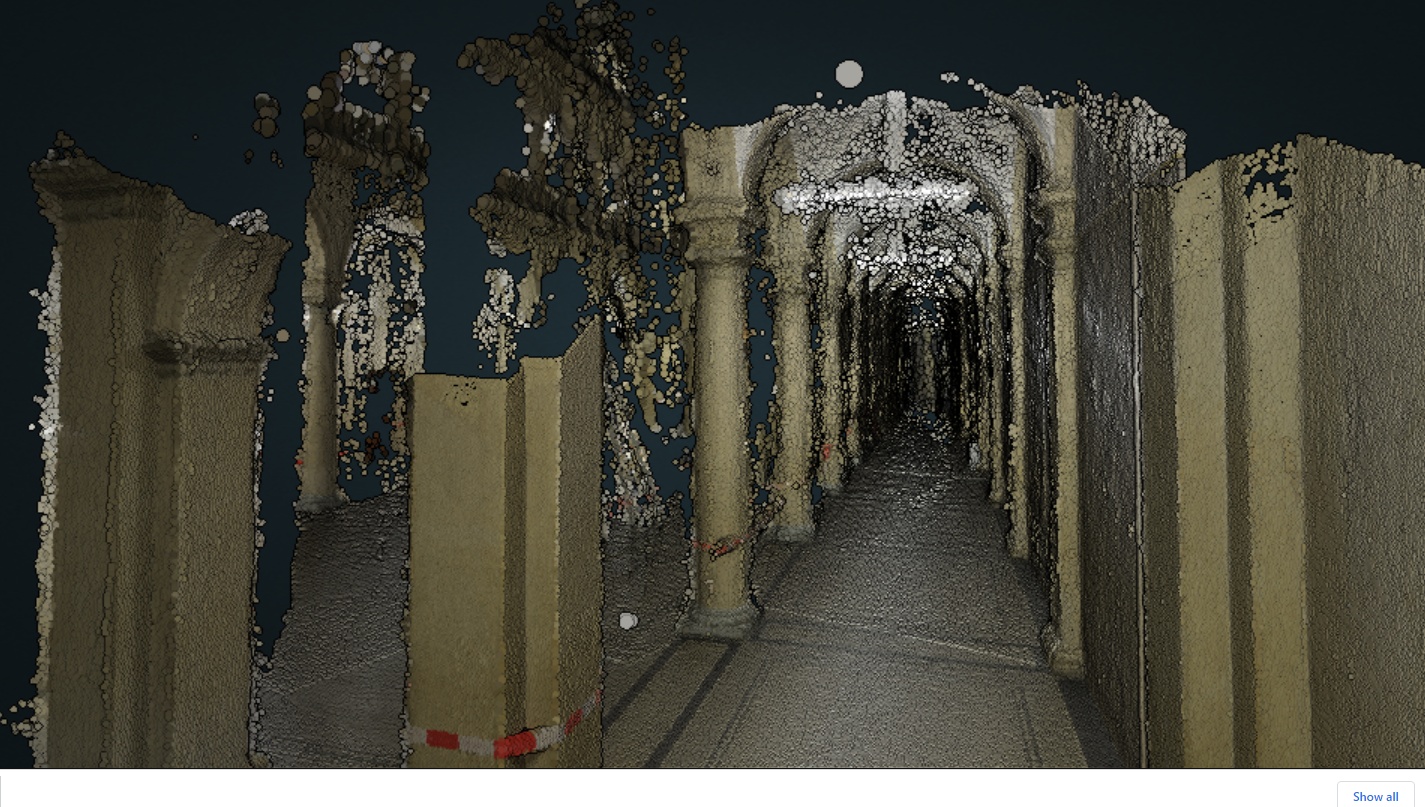}\\
    \includegraphics[trim={0 5em 0 10em},clip,width=0.48\textwidth]{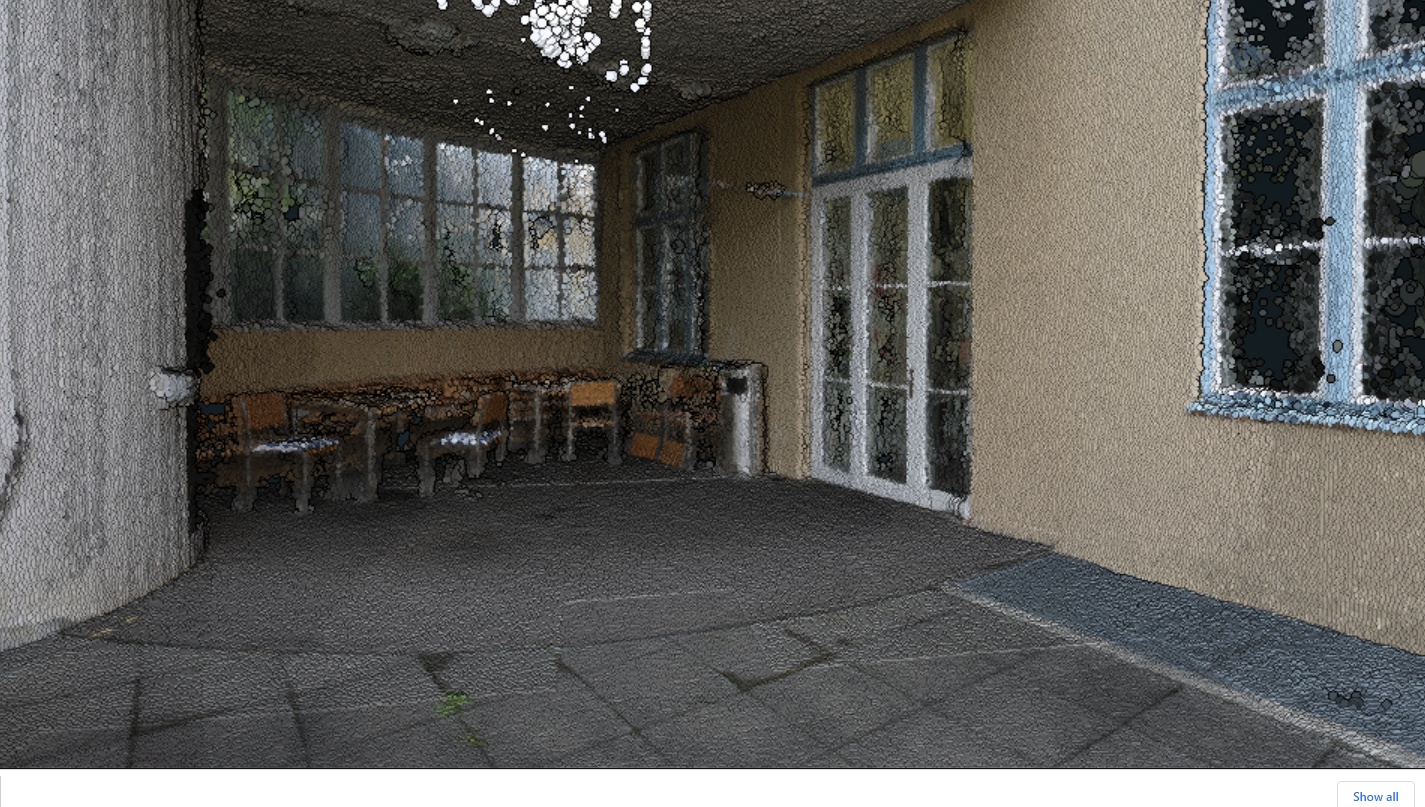}
    \includegraphics[trim={0 5em 0 10em},clip,width=0.48\textwidth]{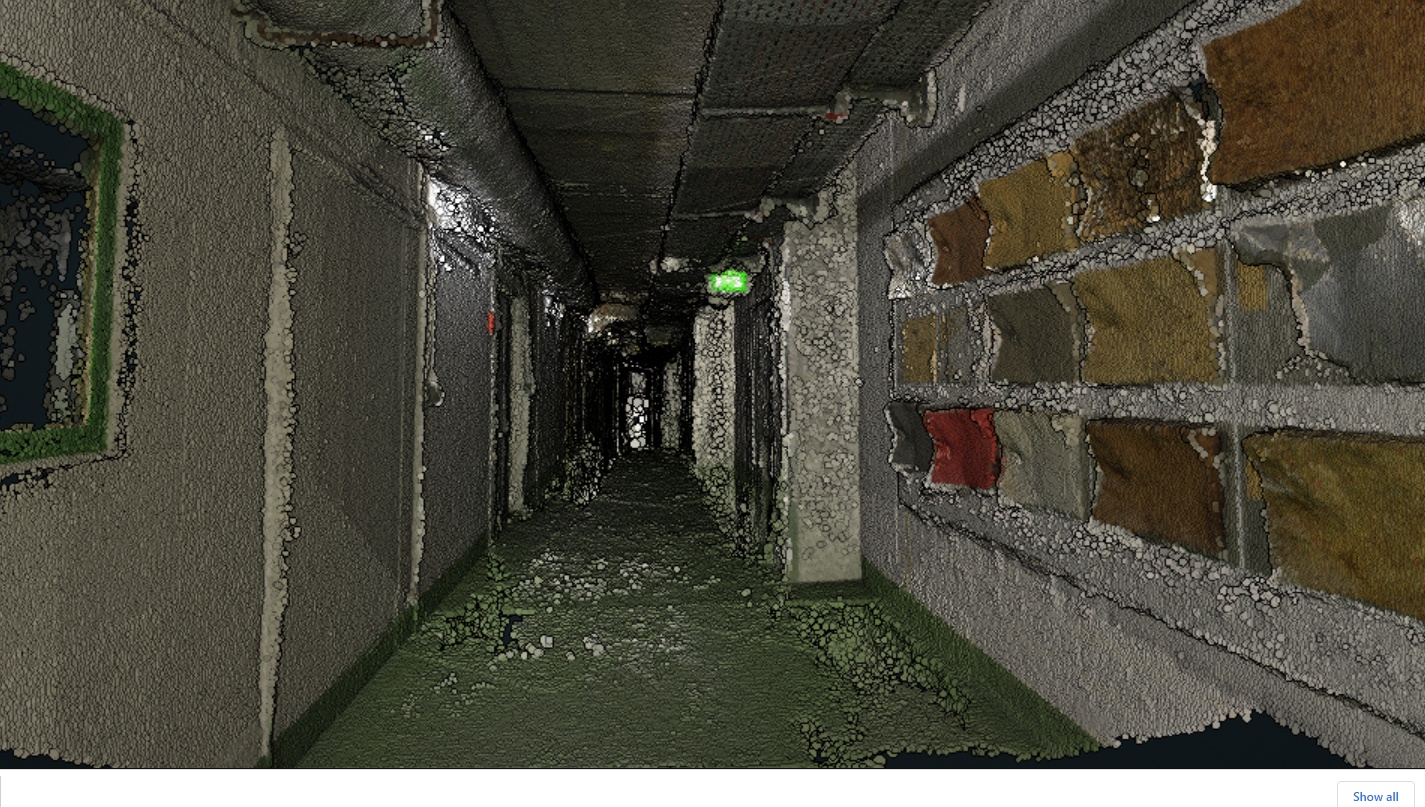}\\
    \includegraphics[trim={0 5em 0 10em},clip,width=0.48\textwidth]{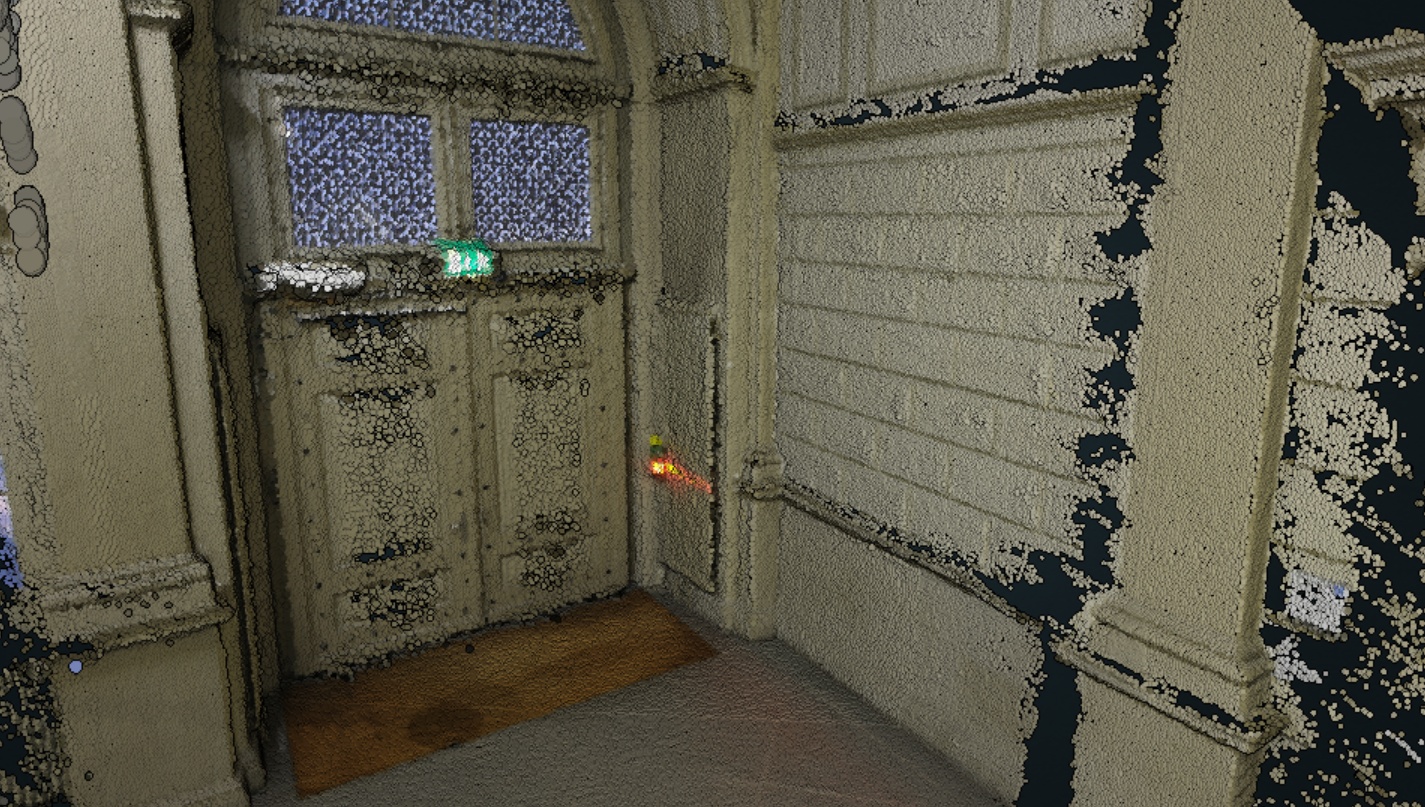}
    \includegraphics[trim={0 5em 0 10em},clip,width=0.48\textwidth]{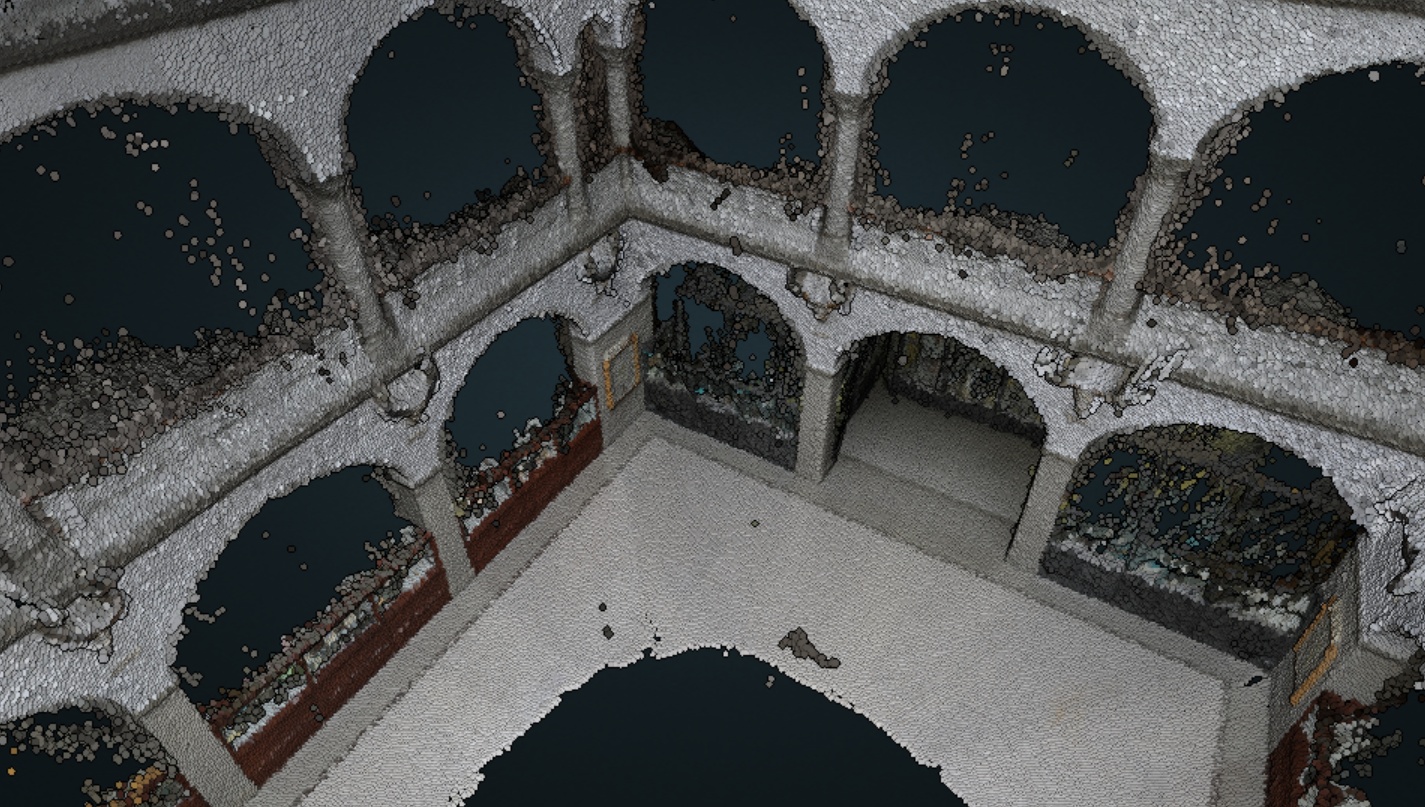}\\
    \includegraphics[trim={0 5em 0 10em},clip,width=0.48\textwidth]{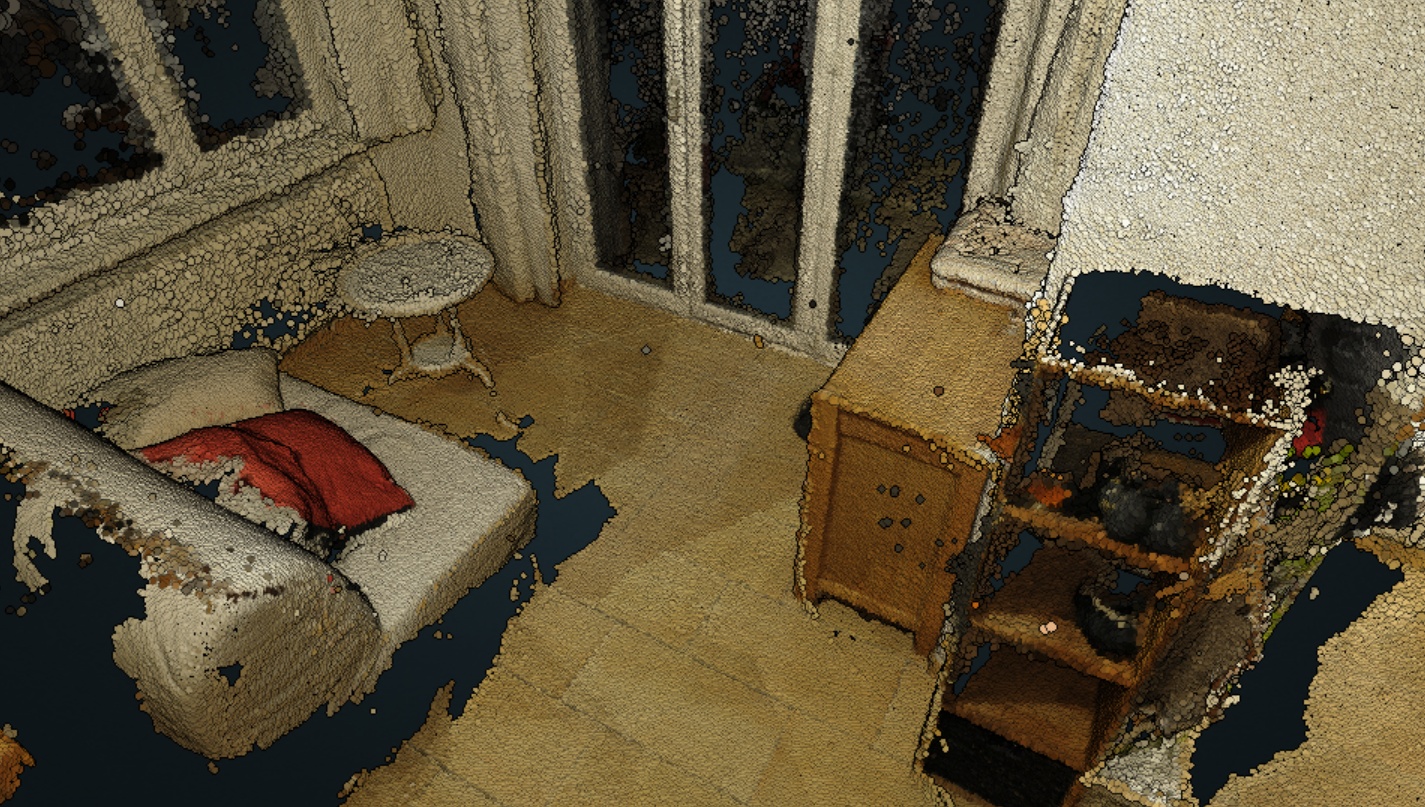}
    \includegraphics[trim={0 5em 0 10em},clip,width=0.48\textwidth]{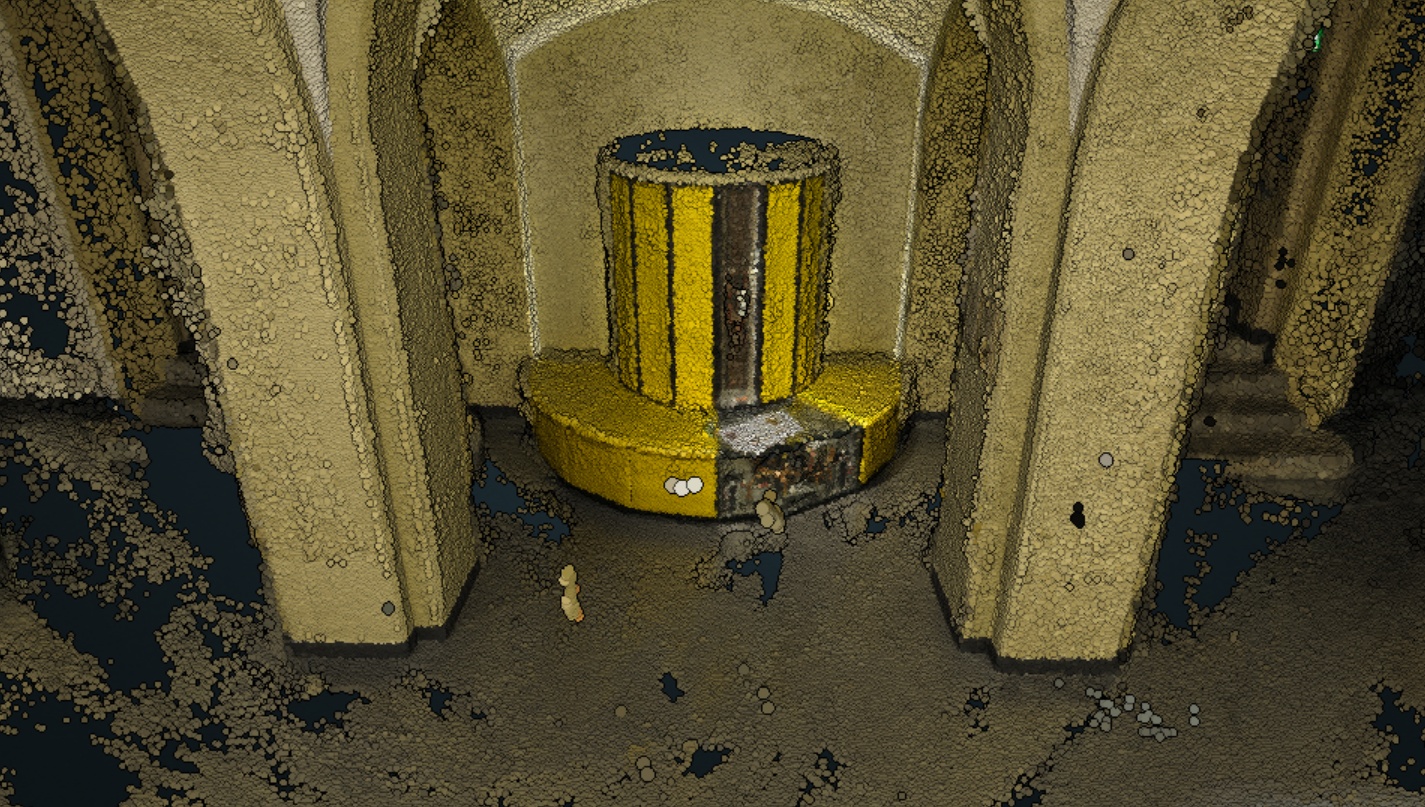}\\
    \caption{ \textbf{Additional Qualitative Results for ETH3D High-Res~\cite{schoeps2017eth3d} Train set (top two rows) and Test set (bottom two rows) for $\text{Ours}_{lr}$}.}
    \label{fig:eth_lr_train} 
\end{figure*}

%% file: figures/supp/tnt.tex
\begin{figure*}[t]
    \centering
    \includegraphics[width=0.48\textwidth]{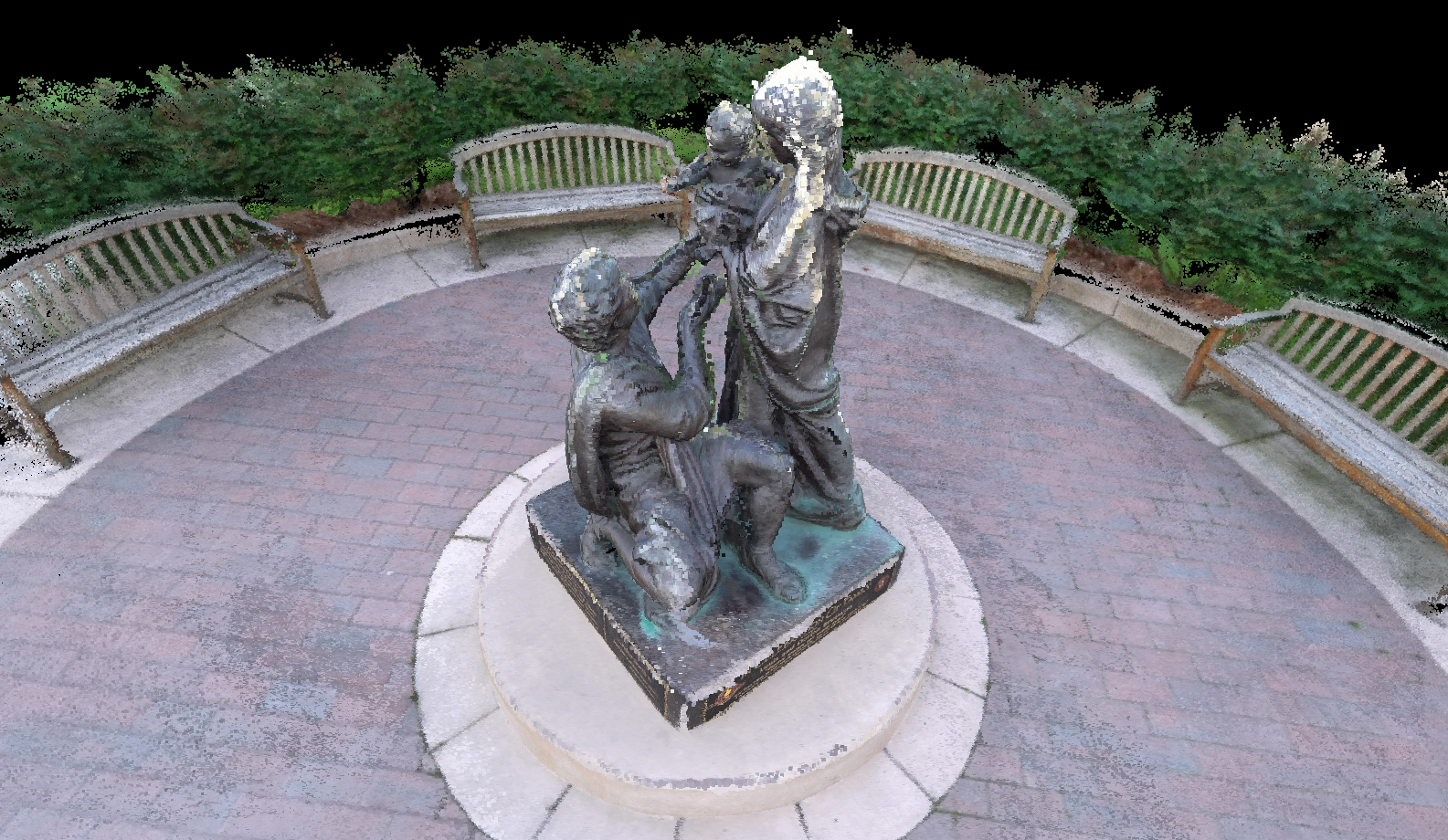}
    \includegraphics[width=0.48\textwidth]{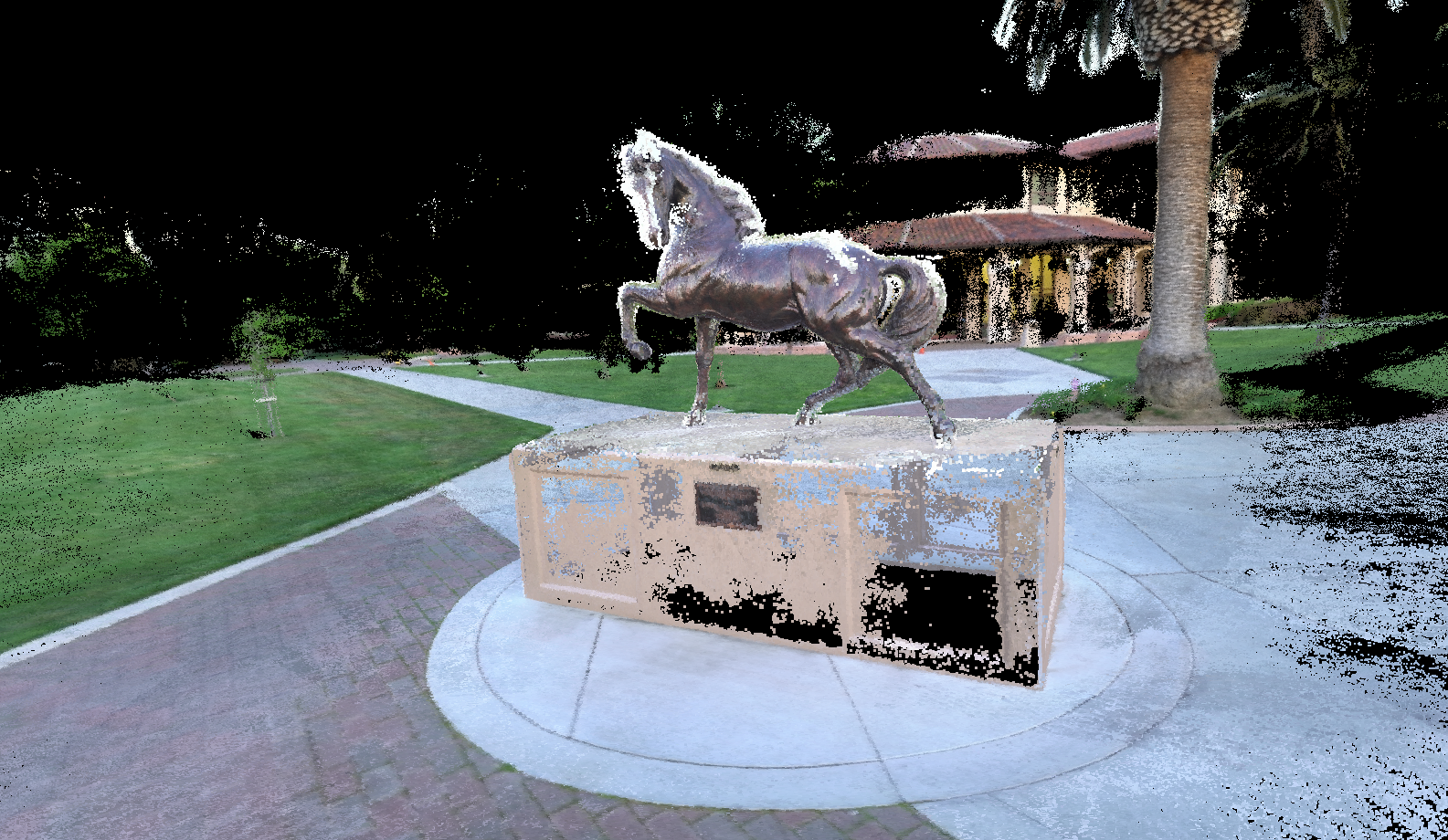}\\
    \includegraphics[width=0.48\textwidth]{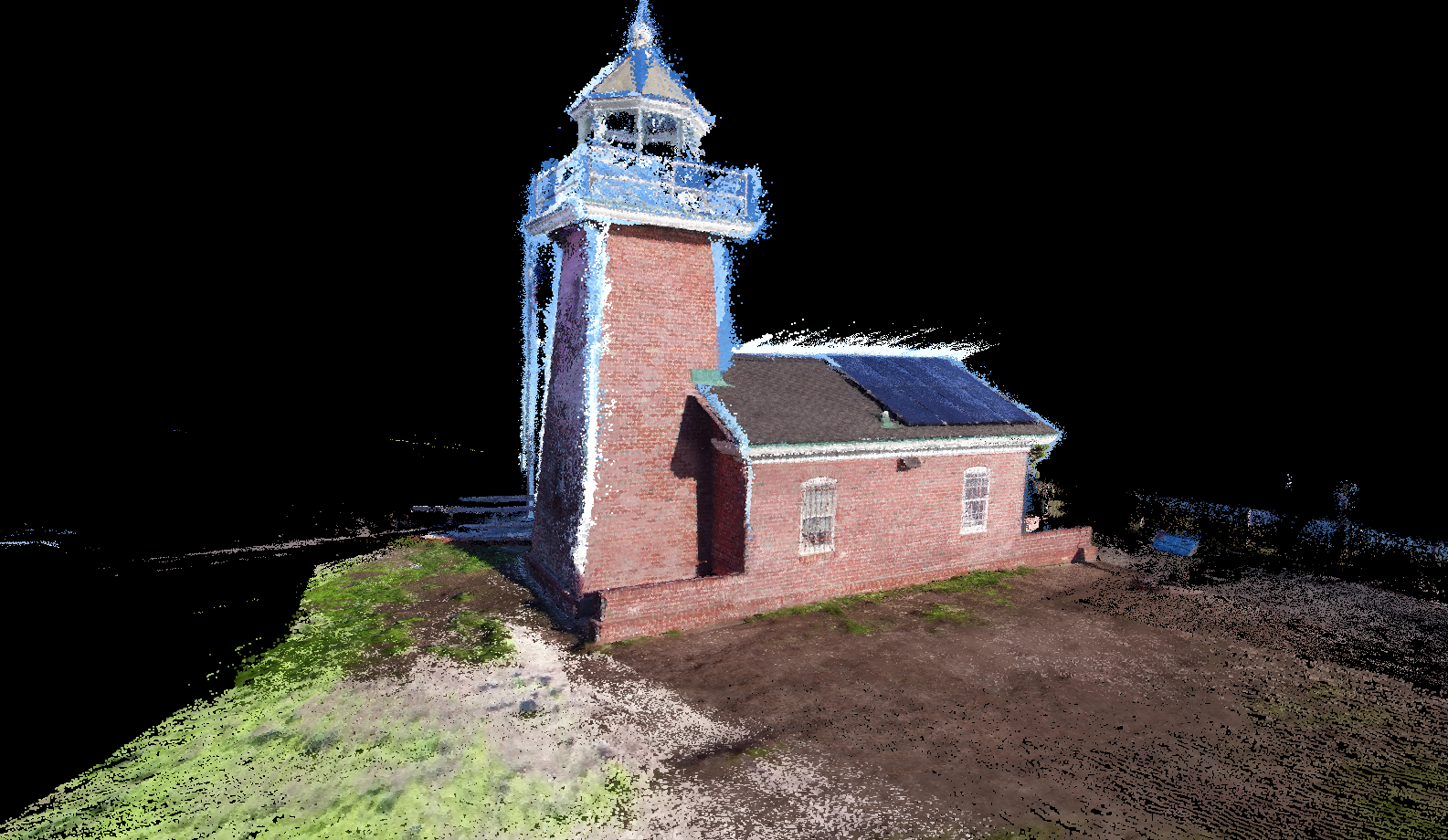}
    \includegraphics[width=0.48\textwidth]{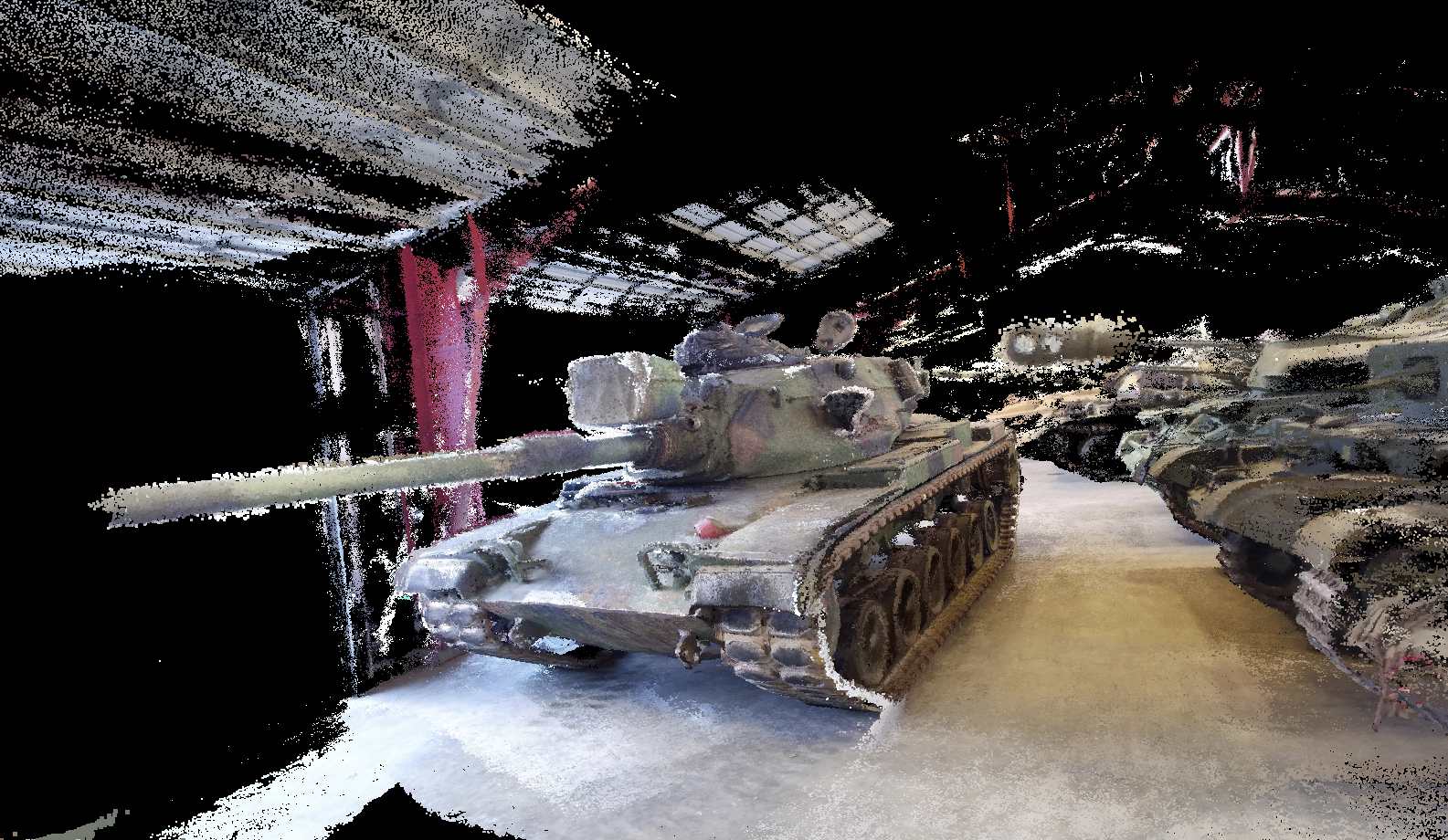}\\
    \caption{ \textbf{Additional Qualitative Results for Tanks and Temples~\cite{Knapitsch2017tanks} Intermediate set.} }
    \label{fig:tnt_intermediate} 
\end{figure*}

%% file: figures/supp/tnt_advanced.tex
\begin{figure*}[t]
    \centering
    \includegraphics[width=0.48\textwidth]{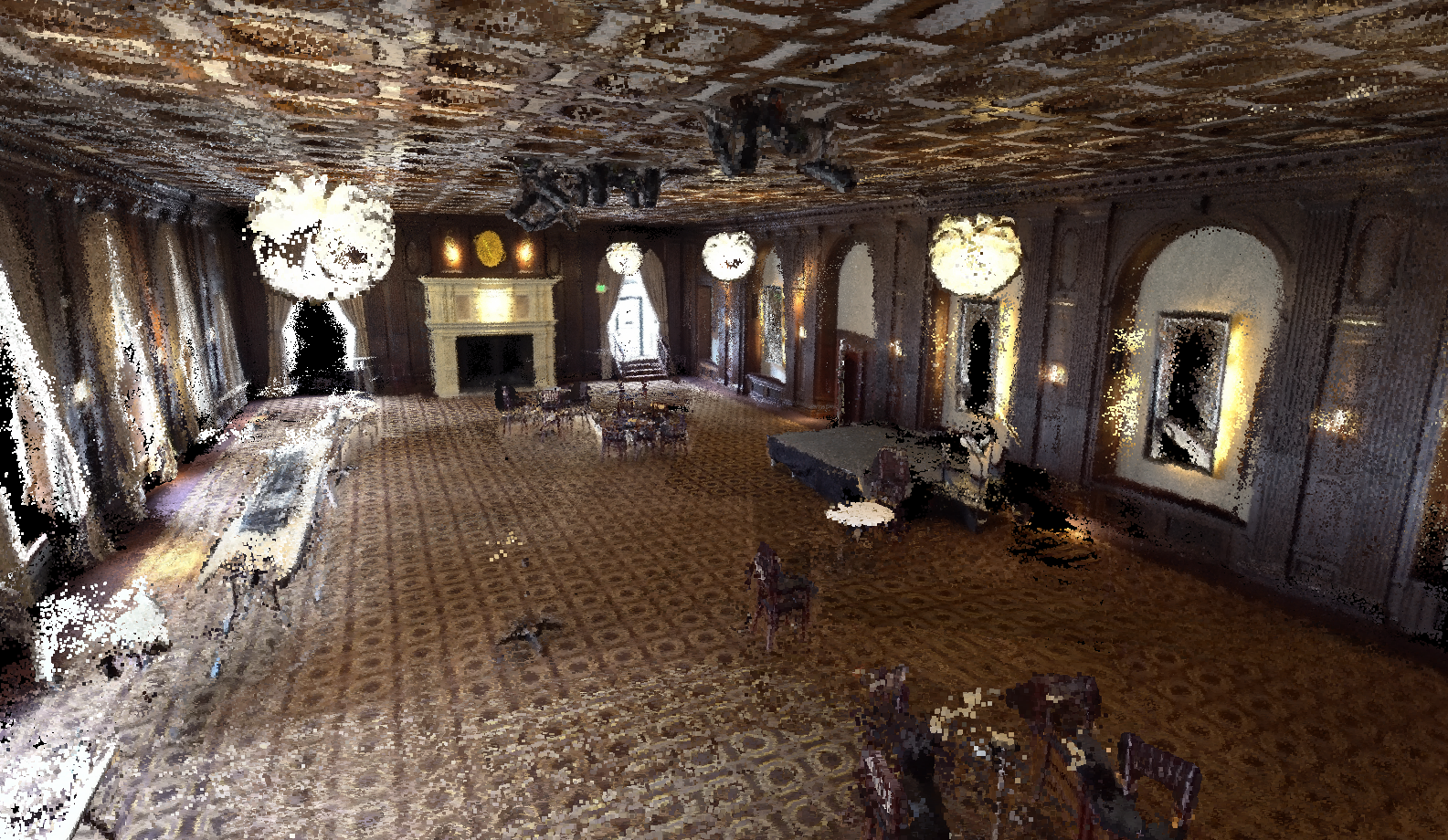}
    \includegraphics[width=0.48\textwidth]{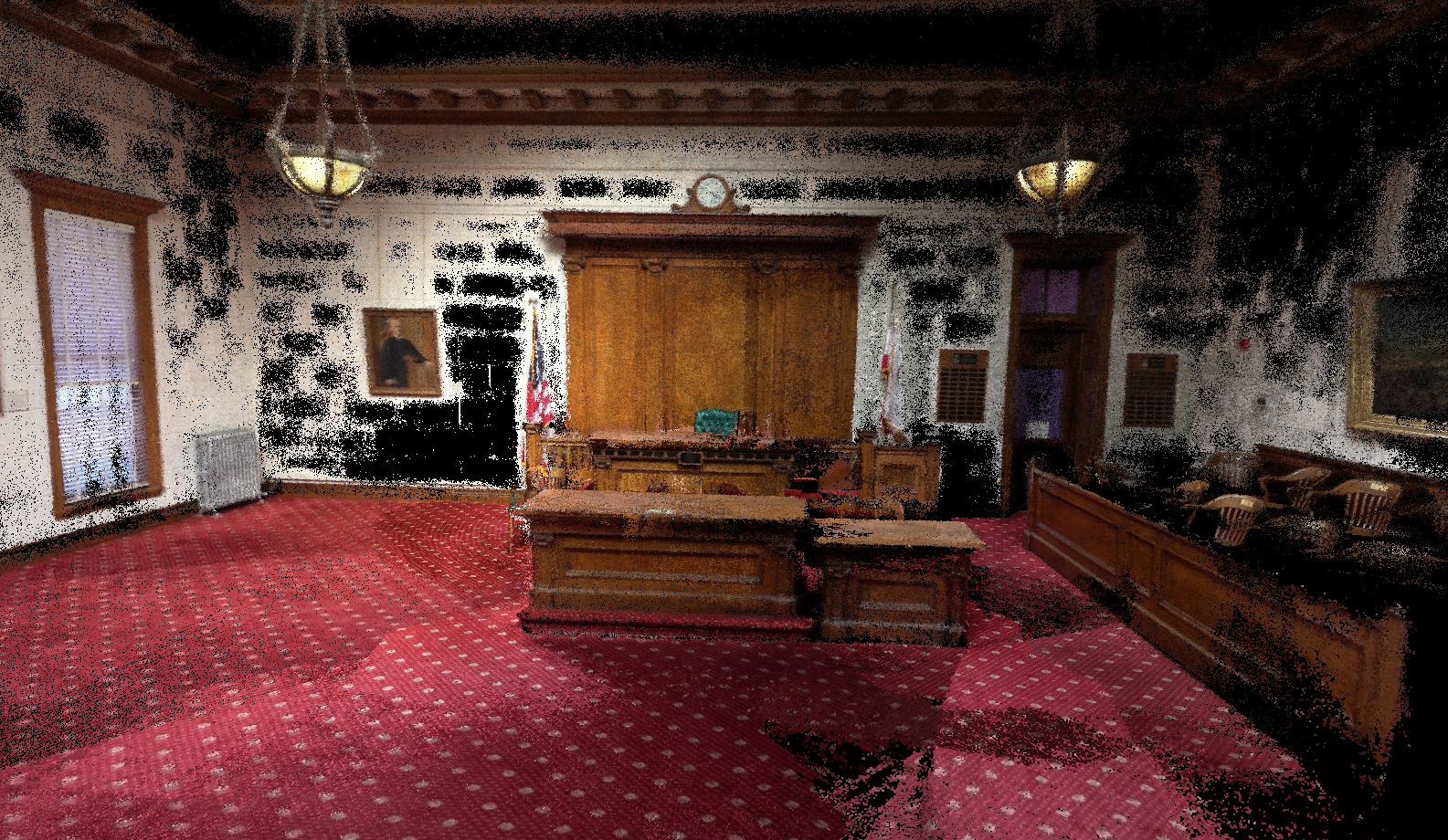}\\
    \includegraphics[width=0.48\textwidth]{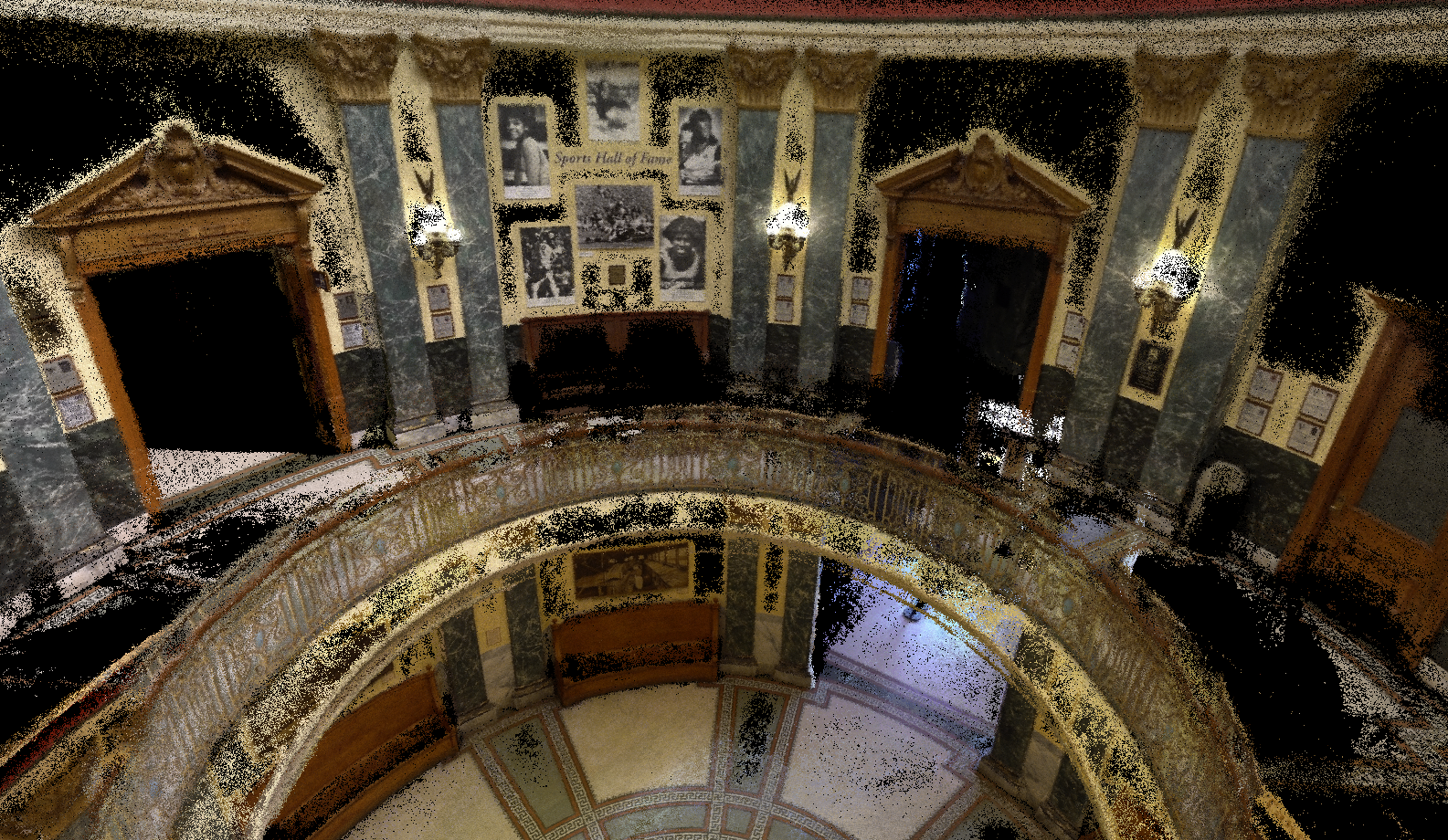}
    \includegraphics[width=0.48\textwidth]{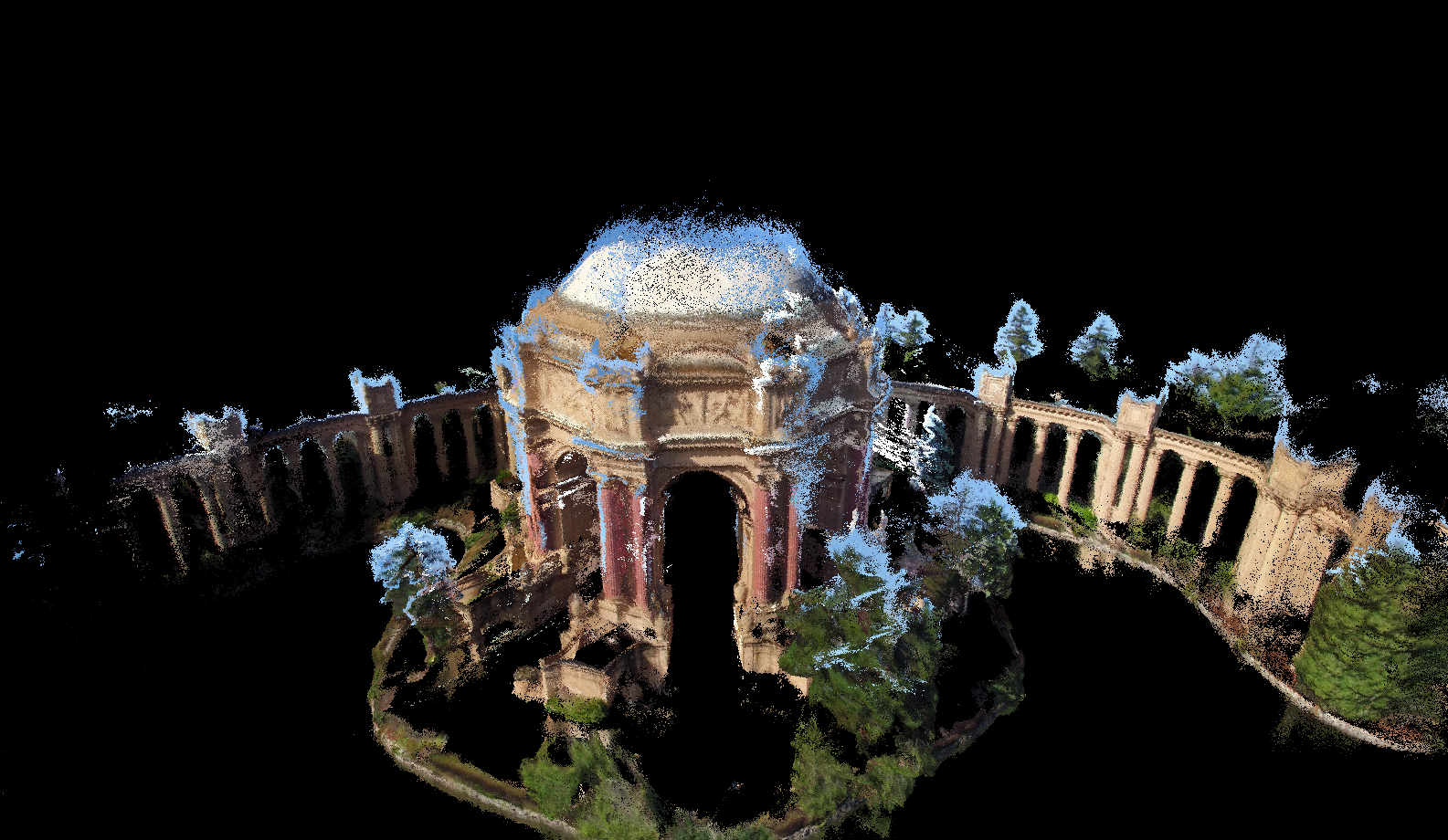}\\
    \caption{ \textbf{Additional Qualitative Results for Tanks and Temples~\cite{Knapitsch2017tanks} Advanced set.} }
    \label{fig:tnt_advanced} 
\end{figure*}

%% file: tex/conclusion.tex
\section{Conclusion}
We have proposed a learning-based PatchMatch MVS method that benefits from 
learned patch coplanarity, geometric consistency and adaptive pixel sampling. Our experiments show that each of these individually contribute to higher performance and together lead to much more complete models while retaining high accuracy, comparing well to state-of-the-art learning and non-learning based methods on challenging benchmarks.  



%% file: tex/acknowledgments.tex
\bmhead{Data Availability Statement}
The data that support the findings of this study are openly available at \url{https://doi.org/10.1109/CVPR42600.2020.00186}~(BlendedMVS)~\cite{yao2020blendedmvs}, \url{https://doi.org/10.1145/3072959.3073599}~(TanksAndTemples)~\cite{Knapitsch2017tanks} and  \url{https://doi.org/10.1109/CVPR.2017.272}~(ETH3D)~\cite{schoeps2017eth3d}.
\bmhead{Acknowledgments}
This research is supported in part by ONR Award N00014-16-1-2007, NSF award IIS 2020227, and gift from Amazon Go Research. 

%% file: main.bbl
\begin{thebibliography}{10}
\providecommand{\url}[1]{\texttt{#1}}
\providecommand{\urlprefix}{URL }
\providecommand{\doi}[1]{https://doi.org/#1}

\bibitem{aanaes2016large}
Aan{\ae}s, H., Jensen, R.R., Vogiatzis, G., Tola, E., Dahl, A.B.: Large-scale
  data for multiple-view stereopsis. International Journal of Computer Vision
  \textbf{120}(2),  153--168 (2016)

\bibitem{anandan1989computational}
Anandan, P.: A computational framework and an algorithm for the measurement of
  visual motion. International Journal of Computer Vision  \textbf{2}(3),
  283--310 (1989)

\bibitem{barnes2009patchmatch}
Barnes, C., Shechtman, E., Finkelstein, A., Goldman, D.B.: Patchmatch: A
  randomized correspondence algorithm for structural image editing

\bibitem{bleyer2011patchmatch}
Bleyer, M., Rhemann, C., Rother, C.: Patchmatch stereo-stereo matching with
  slanted support windows.

\bibitem{cheng2019learning}
Cheng, X., Wang, P., Yang, R.: Learning depth with convolutional spatial
  propagation network. IEEE transactions on pattern analysis and machine
  intelligence  \textbf{42}(10),  2361--2379 (2019)

\bibitem{furukawa2015_mvsTutorial}
Furukawa, Y., Hern{\'a}ndez, C.: Multi-view stereo: A tutorial. Found. Trends
  Comput. Graph. Vis.  \textbf{9},  1--148 (2015)

\bibitem{furukawa2009accurate}
Furukawa, Y., Ponce, J.: Accurate camera calibration from multi-view stereo and
  bundle adjustment. International Journal of Computer Vision  \textbf{84}(3),
  257--268 (2009)

\bibitem{galliani2015massively}
Galliani, S., Lasinger, K., Schindler, K.: Massively parallel multiview
  stereopsis by surface normal diffusion. In: Proceedings of the IEEE
  International Conference on Computer Vision. pp. 873--881 (2015)

\bibitem{golparvar2011monitoring}
Golparvar-Fard, M., Pena-Mora, F., Savarese, S.: Monitoring changes of 3d
  building elements from unordered photo collections. In: 2011 IEEE
  International Conference on Computer Vision Workshops (ICCV Workshops). pp.
  249--256. IEEE (2011)

\bibitem{gu2020cascade}
Gu, X., Fan, Z., Zhu, S., Dai, Z., Tan, F., Tan, P.: Cascade cost volume for
  high-resolution multi-view stereo and stereo matching. In: Proceedings of the
  IEEE/CVF Conference on Computer Vision and Pattern Recognition (CVPR) (June
  2020)

\bibitem{hannah1974computer}
Hannah, M.J.: Computer matching of areas in stereo images. Stanford University
  (1974)

\bibitem{Knapitsch2017tanks}
Knapitsch, A., Park, J., Zhou, Q.Y., Koltun, V.: Tanks and temples:
  Benchmarking large-scale scene reconstruction. ACM Transactions on Graphics
  \textbf{36}(4) (2017)

\bibitem{kuhn2020deepc}
Kuhn, A., Sormann, C., Rossi, M., Erdler, O., Fraundorfer, F.: Deepc-mvs: Deep
  confidence prediction for multi-view stereo reconstruction. In: 2020
  International Conference on 3D Vision (3DV). pp. 404--413. IEEE (2020)

\bibitem{lee2021patchmatchrl}
Lee, J.Y., DeGol, J., Zou, C., Hoiem, D.: Patchmatch-rl: Deep mvs with
  pixelwise depth, normal, and visibility. In: Proceedings of the IEEE/CVF
  International Conference on Computer Vision (October 2021)

\bibitem{lucasKanade1981_stereo}
Lucas, B.D., Kanade, T.: An iterative image registration technique with an
  application to stereo vision. In: Proc. International Joint Conference on
  Artificial Intelligence (IJCAI). p. 674–679 (1981)

\bibitem{luo2020attention}
Luo, K., Guan, T., Ju, L., Wang, Y., Chen, Z., Luo, Y.: Attention-aware
  multi-view stereo. In: Proceedings of the IEEE/CVF Conference on Computer
  Vision and Pattern Recognition. pp. 1590--1599 (2020)

\bibitem{ma2021eppmvsnet}
Ma, X., Gong, Y., Wang, Q., Huang, J., Chen, L., Yu, F.: Epp-mvsnet:
  Epipolar-assembling based depth prediction for multi-view stereo. In:
  Proceedings of the IEEE/CVF International Conference on Computer Vision
  (ICCV). pp. 5732--5740 (October 2021)

\bibitem{NEURIPS2019_9015}
Paszke, A., Gross, S., Massa, F., Lerer, A., Bradbury, J., Chanan, G., Killeen,
  T., Lin, Z., Gimelshein, N., Antiga, L., Desmaison, A., Kopf, A., Yang, E.,
  DeVito, Z., Raison, M., Tejani, A., Chilamkurthy, S., Steiner, B., Fang, L.,
  Bai, J., Chintala, S.: Pytorch: An imperative style, high-performance deep
  learning library. In: Wallach, H., Larochelle, H., Beygelzimer, A.,
  d\textquotesingle Alch\'{e}-Buc, F., Fox, E., Garnett, R. (eds.) Advances in
  Neural Information Processing Systems 32, pp. 8024--8035. Curran Associates,
  Inc. (2019)

\bibitem{prokopetc2019towards}
Prokopetc, K., Dupont, R.: Towards dense 3d reconstruction for mixed reality in
  healthcare: Classical multi-view stereo vs deep learning. In: Proceedings of
  the IEEE/CVF International Conference on Computer Vision Workshops. pp.~0--0
  (2019)

\bibitem{rebecq2018emvs}
Rebecq, H., Gallego, G., Mueggler, E., Scaramuzza, D.: Emvs: Event-based
  multi-view stereo—3d reconstruction with an event camera in real-time.
  International Journal of Computer Vision  \textbf{126}(12),  1394--1414
  (2018)

\bibitem{romanoni2019tapa}
Romanoni, A., Matteucci, M.: Tapa-mvs: Textureless-aware patchmatch multi-view
  stereo. In: Proceedings of the IEEE International Conference on Computer
  Vision. pp. 10413--10422 (2019)

\bibitem{ronneberger2015u}
Ronneberger, O., Fischer, P., Brox, T.: U-net: Convolutional networks for
  biomedical image segmentation. In: International Conference on Medical image
  computing and computer-assisted intervention. pp. 234--241. Springer (2015)

\bibitem{schoenberger2016mvs}
Sch\"{o}nberger, J.L., Zheng, E., Pollefeys, M., Frahm, J.M.: {Pixelwise View
  Selection for Unstructured Multi-View Stereo}. In: European Conference on
  Computer Vision (ECCV) (2016)

\bibitem{schoeps2017eth3d}
Sch\"ops, T., Sch\"onberger, J.L., Galliani, S., Sattler, T., Schindler, K.,
  Pollefeys, M., Geiger, A.: A multi-view stereo benchmark with high-resolution
  images and multi-camera videos. In: Conference on Computer Vision and Pattern
  Recognition (CVPR) (2017)

\bibitem{wang2020patchmatchnet}
Wang, F., Galliani, S., Vogel, C., Speciale, P., Pollefeys, M.: Patchmatchnet:
  Learned multi-view patchmatch stereo (2020)

\bibitem{wang2020mesh}
Wang, Y., Guan, T., Chen, Z., Luo, Y., Luo, K., Ju, L.: Mesh-guided multi-view
  stereo with pyramid architecture. In: Proceedings of the IEEE/CVF Conference
  on Computer Vision and Pattern Recognition. pp. 2039--2048 (2020)

\bibitem{xu2019multi}
Xu, Q., Tao, W.: Multi-scale geometric consistency guided multi-view stereo.
  In: Proceedings of the IEEE Conference on Computer Vision and Pattern
  Recognition. pp. 5483--5492 (2019)

\bibitem{Xu2020ACMP}
Xu, Q., Tao, W.: Planar prior assisted patchmatch multi-view stereo. AAAI
  Conference on Artificial Intelligence (AAAI)  (2020)

\bibitem{xu2020marmvs}
Xu, Z., Liu, Y., Shi, X., Wang, Y., Zheng, Y.: Marmvs: Matching ambiguity
  reduced multiple view stereo for efficient large scale scene reconstruction.
  In: Proceedings of the IEEE/CVF Conference on Computer Vision and Pattern
  Recognition (CVPR) (June 2020)

\bibitem{yang2013image}
Yang, M.D., Chao, C.F., Huang, K.S., Lu, L.Y., Chen, Y.P.: Image-based 3d scene
  reconstruction and exploration in augmented reality. Automation in
  Construction  \textbf{33},  48--60 (2013)

\bibitem{yao2018mvsnet}
Yao, Y., Luo, Z., Li, S., Fang, T., Quan, L.: Mvsnet: Depth inference for
  unstructured multi-view stereo. In: Proceedings of the European Conference on
  Computer Vision (ECCV). pp. 767--783 (2018)

\bibitem{yao2019recurrent}
Yao, Y., Luo, Z., Li, S., Shen, T., Fang, T., Quan, L.: Recurrent mvsnet for
  high-resolution multi-view stereo depth inference. In: Proceedings of the
  IEEE Conference on Computer Vision and Pattern Recognition. pp. 5525--5534
  (2019)

\bibitem{yao2020blendedmvs}
Yao, Y., Luo, Z., Li, S., Zhang, J., Ren, Y., Zhou, L., Fang, T., Quan, L.:
  Blendedmvs: A large-scale dataset for generalized multi-view stereo networks.
  Computer Vision and Pattern Recognition (CVPR)  (2020)

\bibitem{zhang2020visibility}
Zhang, J., Yao, Y., Li, S., Luo, Z., Fang, T.: Visibility-aware multi-view
  stereo network. British Machine Vision Conference (BMVC)  (2020)

\bibitem{zheng2014patchmatch}
Zheng, E., Dunn, E., Jojic, V., Frahm, J.M.: Patchmatch based joint view
  selection and depthmap estimation. In: Proceedings of the IEEE Conference on
  Computer Vision and Pattern Recognition. pp. 1510--1517 (2014)

\end{thebibliography}
